\documentclass[runningheads]{llncs}
\usepackage{amsmath,amssymb,amsfonts}
\usepackage{graphicx}
\usepackage{xcolor}
\usepackage{colortbl}
\usepackage{hhline}
\usepackage{booktabs}
\usepackage[sectionbib]{natbib}
\usepackage{pgfplots}
\usepackage{multirow}
\usepackage{url}
\usepackage{comment}

\usepackage{subcaption}
\definecolor{my_red}{RGB}{150,0,0}
\definecolor{my_green}{RGB}{0,150,0}
\definecolor{my_grey}{RGB}{50,50,50}
\definecolor{my_grey2}{RGB}{100,100,100}

\definecolor{my_remove_red}{RGB}{250,130,130}
\definecolor{my_creation_blue}{RGB}{0,0,200}
\definecolor{my_creation_orange}{RGB}{255,100,0}
\newcommand{\remMoritz}[1]{{\leavevmode\color{my_remove_red}MK remove:#1}}

\newcommand{\comMoritz}[1]{{\leavevmode\color{my_creation_orange}MK:#1}}
\newcommand{\comMoritzOld}[1]{{\leavevmode\color{my_creation_orange}MK:#1}}

\newcommand{\skpcandidate}[1]{{\leavevmode\color{my_grey2}Skip?: #1}}

\newcommand{\com}[1]{{\leavevmode\color{blue}COM: #1}}
\newcommand{\res}[1]{{\leavevmode\color{green}RES: #1}}
\newcommand{\tod}[1]{{\leavevmode\color{red}TOD: #1}}
\newcommand{\raus}[1]{{\leavevmode\color{gray}RAUS: #1}}

\newcommand{\skp}[1]{}
\newcommand{\drin}[1]{#1}

\newcommand{\A}[1]{{\leavevmode\color{brown}RDT: #1}}
\newcommand{\rausA}[1]{{\leavevmode\color{gray}RAUS RDT: #1}}
\newcommand{\B}[1]{{\leavevmode\color{blue}XGB: #1}}
\newcommand{\comds}[1]{\textcolor{red}{#1}}
\newcommand{\rev}[1]{{\leavevmode\color{black}#1}}

 \renewcommand{\com}[1]{}
 \renewcommand{\comds}[1]{}
 \renewcommand{\res}[1]{}
 \renewcommand{\tod}[1]{}
 
 \renewcommand{\raus}[1]{}
 \renewcommand{\com}[1]{}
 \renewcommand{\com}[1]{}
 \renewcommand{\tod}[1]{}
 \renewcommand{\skp}[1]{}
 \renewcommand{\skpcandidate}[1]{}
 \renewcommand{\remMoritz}[1]{}

 \renewcommand{\comMoritz}[1]{}
 \renewcommand{\comMoritzOld}[1]{}
 \renewcommand{\skpcandidate}[1]{}
 \renewcommand{\A}[1]{}
 \renewcommand{\rausA}[1]{}
 \renewcommand{\B}[1]{}

\DeclareMathOperator{\tp}{\ensuremath \textit{tp}}
\DeclareMathOperator{\fp}{\ensuremath \textit{fp}}

\DeclareMathOperator{\fn}{\ensuremath \textit{fn}}

\DeclareMathOperator*{\argmax}{arg\,max}
\DeclareMathOperator*{\sigmoid}{sigmoid}
\DeclareMathOperator*{\argmin}{arg\,min}

\newcommand{\vektor}[1]{\ensuremath{\mathrm{\mathbf{#1}}}} 

\newcommand{\vx}{\vektor{x}}
\newcommand{\vy}{\vektor{y}}
\newcommand{\vp}{\vektor{p}}

\newcommand{\given}{\, | \,}

\newcommand{\ds}[1]{\textsc{#1}}
\newcommand{\ap}[1]{\textsl{#1}}

\begin{document}
%
\title{
Tree-Based Dynamic Classifier Chains}
%
%
\author{Eneldo Loza Menc\'ia \and Moritz Kulessa \and \\ Simon Bohlender \and Johannes F\"urnkranz}
\authorrunning{E. Loza Menc\'ia et al.}
%
\institute{Knowledge Engineering Group,  Technische Universit\"at Darmstadt,
Germany \and
Computational Analytics Group, Johannes Kepler Universit\"at Linz, Austria
\\
\email{research@eneldo.net
\and
moritz.kulessa@gmail.com
\and
simon.bohlender@gmail.com
\and
juffi@faw.jku.at
}
}
\maketitle              
\begin{abstract}
Classifier chains are an effective technique for modeling label dependencies in multi-label classification.
However, the method requires a fixed, static order of the labels. While in theory, any order is sufficient,
in practice, this order has a substantial impact on the quality of the final prediction.
Dynamic classifier chains denote the idea that for each instance to classify, the order in which the labels are predicted is dynamically chosen. The complexity of a na\"ive implementation of such an approach is prohibitive, because it would require to train a sequence of classifiers for every possible permutation of the labels.
To tackle this problem efficiently, we propose a new approach based on random decision trees which can dynamically select the label ordering for each prediction.
We show empirically that a dynamic selection of the next label improves over the use of a static ordering under an otherwise unchanged random decision tree model. 
%
In addition, we also demonstrate an alternative approach based on extreme gradient boosted trees, which allows for a more target-oriented training of dynamic classifier chains.
Our results show that this variant outperforms random decision trees and other tree-based multi-label classification methods. More importantly, the dynamic selection strategy allows to considerably speed up training and prediction.


\end{abstract}
\tod{
- upload RDT Repo
- bei arxiv hochladen (vorher überprüfen, ob es laut MLJ Richtlinien ok ist)
- Graphen für alle Datensätze erstellen, insbesondere predictive performance. Hier könnte man auch ex. based recall und precision hinzufügen, um Erklärungen zu haben.
Nice to have:
- XDCC Experimente neu machen um die C++ Zeiten zu messen
}
\section{Introduction }
\label{sec:intro}

Contrary to multi-class classification, where only one class label is expected to be associated to an example,
multi-label classification (MLC) is the task of assigning a subset of all possible labels to an example \citep{Multilabel-Overview}.
A well known example for such a problem is automatic text categorization where the goal is to assign a set of relevant categories to a document \citep{schapire00boostexter}. Most of the current approaches address these problems with problem transformation methods where the problem is split up into multiple smaller subtasks which are solved independently.
\emph{Binary Relevance} (BR) \rev{\citep{joachims98svmtext,godbole04stacking,Scene-Data}}
is the best known example, which decomposes the multi-label problem into a set of $N$ independent binary classification problems, one for each label.
However, this decomposition ignores possible dependencies between the labels.

\emph{Classifier chains} (CC) solve this problem by learning one model for each label, but taking the predictions of the previous models along a predetermined sequence of the labels into account \citep{cc,cc-review}.
To this end, the models are arranged along a predefined chain where each model passes its own and all previously predicted labels on to the next model in the chain, which incorporates them as additional input features.
It can be shown that CC are able to capture local as well as global dependencies, and that these are crucial for minimizing non-deomposable loss functions, which cannot be reduced to the marginal label errors \citep{dembczynski12PCCdependence}.

However, while these theoretical results are independent of the selected chain order, in practice the performance of CC highly depends on the order of the labels along the chain. There are several reasons for this, among them, e.g., the propagation of label errors through the chains: if the first label in the chain is incorrectly selected, this error may affect all subsequent predictions \citep{senge2014problem}.
Finding a good sequence is a non-trivial task because
(i) the number of possible sequences to consider grows exponentially with the number of labels, and (ii)
local dependencies may make it necessary to consider different chains for different instances.
For example, 
it is arguable easier to detect first a car and then infer its headlights in an image taken at daylight, whereas it is easier to first detect the lights and from that deduce the car in a night image.
Moreover, methods which try to explore different orderings are usually computationally expensive so that the most frequently selected option is to pick a random ordering.

The assumption in this work is that the ordering in which labels should be predicted in order to obtain the best performance highly depends on the specific context, namely the test instance at hand.
\emph{Dynamic chaining} addresses this problem  of how to dynamically choose an appropriate ordering for individual instances instead of the entire datasets. The main problem that needs to be solved is that, if the label prediction order can be dynamically selected at prediction time, we need to prepare the classifier for an exponential number of potential label orderings. A na\"ive adaptation of CC for this setting, would therefore have to train up to $N!$ different chains, which is clearly infeasible. The contribution of this work is to investigate tree-based ensembles for an efficient solution of this task.


Our first contribution is the adaptation of \emph{random decision trees} (RDT) \citep{rdt_ml} for the purpose of constructing dynamic chains.
In contrast to the common induction of decision trees or to random forests, RDTs 
do not optimize a heuristic splitting criterion, but instead select the splits at the inner nodes randomly.
We adapt RDTs to dynamic chains by allowing
tests on the labels at the inner nodes, which can be turned on or off.
This has the advantage that the objective can easily be changed during prediction without the need for modifying the trees.
In an iterative process, the learned RDTs are repeatedly queried to determine the next most certain (positive or negative) label, which is then
added to the input features, and the respective label tests in the RDTs are turned on.
Furthermore they are fast to train and can achieve competitive and robust performance without optimizing any objective function~\citep{rdt_hash}.
Most importantly, RDTs allow us to embed static and dynamic chains in exactly the same trees, so that we can compare the two approaches in a controlled environment, with otherwise identical models.
The results of this experimental evaluation confirm that dynamic classification chains clearly outperform static orderings.

However, despite the appealing simplicity and flexibility of RDT, this comes at the expense of predictive performance in comparison to other tree-based MLC models, since RDTs
are not trained in order to optimize a particular measure.
We therefore further adapt \emph{extreme gradient boosted trees} (XGBoost)
\citep{chen2016xgboost},
a highly optimized and efficient tree induction method,
to the DCC setting.
The resulting classifier, XDCC,\footnote{Publicly available at \url{https://github.com/keelm/XDCC}.\tod{...Both algorithms available at...}} thus integrates DCC into the extreme boosting structure of gradient boosted trees.
XDCC's optimization goal in each boosting round is to predict only one single positive label for which it is the most certain.
This label can be different for each training instance and depends on the given data, label dependencies, and previous predictions for the instance at hand.
The information about the predicted labels is carried over to subsequent rounds. %
A key advantage of the proposed approach is the reduced run time
resulting from the fact that we do not need to predict the entire chain of labels ($N$ predictors), but only the actually relevant labels for each instance, which is typically much smaller (usually below 10).
Hence, only few rounds are potentially enough if only the positive labels are predicted, whereas CC-based approaches have to still make predictions for each of the existing labels.

In summary, our contributions are the following:
 \begin{itemize}

    \item We present a general motivation and a thorough formalization of dynamic classifier chains, and a brief review of previous work in this area (Section~\ref{sec:DCC})

    \item We show how to adapt random decision trees so that static and dynamic classifier chains can be modeled in the same structure, and thus be compared to each other within an otherwise identical model. The results confirm a clear advantage for dynamic over static classifier chains (Section~\ref{sec:dynamicpredictions}).

     \item
     We introduce a multi-label formulation of the XGBoost objective which is much more efficient than the decomposition based XGBoost baselines,  
     and propose a variant of XGBoost which integrates dynamic classifier chains, yielding a versatile and efficient multi-label classifier (Section~\ref{sec:DCC-XGBoost}).
     \item We demonstrate empirically that the XGBoost variant outperforms other tree- and ensemble-based multi-label classifiers, especially regarding the computational costs (Section~\ref{sec:XGBevaluation}).
  \end{itemize}

This paper is based on \citep{mk:DS-18-DCC-RDT} and \citep{mk:DS-20-DCC-XGBoost}. It provides an expanded, unified and definite description of these works, puts a stronger emphasis on the DCC framework,
a more complete discussion of related work, and more detailed as well as several new experimental results.

\section{Multi-label Classification}
\label{sec:multi-label}

This section briefly recapitulates previous work in multi-label classification that is relevant for us and clarifies the notation used throughout this paper.
Extensive overviews of MLC are provided, e.g., by \citet{Multilabel-Overview}, \citet{ml_tsoumakas_mining}, or \citet{zhang2014review}.

\subsection{Problem Definition and Simple Transformation Methods}

Multi-label classification is the task of learning a mapping from instances $X \in \mathcal{X}$ to subsets 
of a finite set of non-exclusive class labels $\mathcal{Y}=\{y_1, \ldots, y_N\}$.
Equivalently, it may be viewed as an instance of multi-target prediction \citep{waegeman19multitarget}, where the task is to predict for a finite set of $N$ unique class labels $\Lambda=\{\lambda_1,\ldots,\lambda_N\}$ whether they are positive (or \emph{relevant}) or negative (or \emph{irrelevant}). Formally, $y_j=1$ if $\lambda_j$ is relevant, and $y_j=0$ if $\lambda_j$ is irrelevant for a given instance. Thus, the training set consists of training examples $\vx_i \in X$ and associated label sets $\vy_i \in \mathcal{Y}=\{0,1\}^N$,  $1 \le i \le M$, which can be represented as matrices $X=(x_{iq}) \in A^{M\times Q}$ and $Y=(y_{ij})\in \{0,1\}^{M\times N}$, where  features $x_{ij}$ can be represented as continuous, categorical or binary values.
An MLC classifier \mbox{$f: \mathcal{X} \rightarrow \mathcal{Y}$}
uses the training set in order to learn the mapping between input features and
output labels. The prediction of $f$ for a test example $\vx$ is a binary vector $\hat\vy=f(\vx)$.


The simplest solution to MLC is \emph{binary relevance} decomposition (BR),
where each label is treated as an independent classification task for which a classifier is trained.
Formally, we learn a function $f_i: \mathcal{X} \to \{0,1\}$ for each different $\vy_i$ that has been observed in the training data.
As a result, each prediction for a label is independent of the predictions of the other labels, which is the main disadvantage of this simple technique.

At the other end of the spectrum is the \emph{label power set} transformation (LP),
which reduces the problem of MLC to a single multi-class classification task, where each possible label combination is encoded as a separate and exclusive class.
By predicting all labels at once, this approach naturally takes label dependencies into account.
In addition to the obvious limitations due to the exponential growth of label combinations, LP does not allow to predict label combinations which have not been seen in the training data.
However, this may also be viewed as an advantage under certain circumstances \citep{senge2014problem}.


\subsection{Classifier Chains}
\label{sec:CC}

Classifier chains \citep{cc,cc-review} overcome the disadvantages of the above-mentioned approaches, because they neither assume full label independence nor full dependence.
As in BR, a set of $N$ binary classifiers is trained, but in order to being able to consider dependencies, the classifiers
are connected along a chain.
Each classifier takes the predictions of all previous ones as additional features \rev{in order to predict the respective label}.
%
More specifically, each $f_j$ is trained on an augmented training set $X^{(j)} = [X,Y_{\cdot,1},\ldots, Y_{\cdot,j-1}]$ \rev{containing the true labels in order} to predict the $j$-th target   $Y_{\cdot,j}$ of $Y$. At prediction time, $f_j$ predicts $\hat{y_j}$ based on previous predictions $\hat{y_1},\ldots,\hat{y}_{j-1}$, i.e.,
$
\hat{y}_j= f_j(\vx,\hat{y_1},\ldots,\hat{y}_{j-1})
$
with $\hat{y}_1=f_1(\vx)$.

\citet{cc_prob} analyzed classifier chains in a probabilistic setting, in which the joint probability of the labels can be estimated via the Bayesian chain rule. In theory, this results in Bayes optimal predictions, independent of the chosen order of the labels.
However, in practice, the resulting \emph{probabilistic classifier chains} (PCC) have a much higher time complexity for finding the label combination with the maximum joint probability, and are thus only feasible for datasets with no more than 15 labels~\citep{cc_prob}.
To tackle this problem beam search \citep{kumar2013beam} or A* search \citep{pcc_a} can be used to perform the inferences, which significantly speeds up the process for determining the most probable label subset.
\citet{pcc_overview} give an overview of inference methods for PCC.
Nevertheless, PCC also rely on a predefined, static chain ordering.



Further research revealed that CC and PCC are able to capture global dependencies as well as  dependencies appearing only locally in the instance space \citep{dembczynski12PCCdependence}.
However, while the decomposition is, in theory, order independent according to the Bayesian chain rule, in practice, the performance of classifier chains depends on the chosen, static ordering of the labels  \citep{daSilva2014distinct}.
Consequently, a variety of techniques have been proposed which aim at determining a good static chain sequence in advance.
For this purpose methods such as genetic algorithms~\citep{cc_label_ordering}, prior statistical analyses and Bayesian networks~\citep{LabelDepRuleLearning,cc_bayesian}, ordering according the difficulty of the single-label problems \citep{kumar2013beam}, formulating it as dynamic programming problem \citep{liu15optimality}, or a double Monte Carlo optimization technique
\citep{read2014efficient}
have been proposed.
Alternatively,
 \citet{cc} already suggested to form an ensemble of different chains, each corresponding to  a different, randomly selected permutation of the labels, a proposal that was later refined by \citet{secc}.

However, creating and maintaining an ensemble of CCs is not always feasible~\citep{cc_label_ordering} and also poses the non-trivial problem of combining multiple, dependent multi-label predictions \citep{ln:MLCaggregation}.
More importantly, static label ordering techniques which use the previous predictions to estimate the next label can practically only tackle dependencies in one direction, which may not be optimal for making predictions.
Indeed, already \citet{LabelDepRuleLearning} found that projecting label dependencies into a sequential ordering is a non-trivial task.
Moreover, especially for tasks with local dependencies, which differ in different parts of the instance space, a static ordering may only be able to capture half of the exploitable dependencies in the worst case.
Taking into consideration such dependency structures require
a dynamic, example-dependent approach of ordering the predictions.
We will return to this question in Section~\ref{sec:dynamicpredictions}.
\tod{possible include \citep{doppa14hc} "only relevant labels are predicted in a sequential manner" and more in the latex comments}

\subsection{Evaluation Measures}
\label{sec:measures}

A large variety of evaluation measures have been proposed in MLC, which differ, e.g., in the importance they attribute to label dependencies. For our purposes,
%
the most interesting ones 
are Hamming accuracy, subset accuracy, and the F1 measure.

\emph{Hamming accuracy} (HA) denotes the accuracy of predicting individual labels averaged over all labels.
\begin{equation}
\text{HA} = \frac1N \sum_{j=1}^N \mathbb{I}\left[y_j = \hat{y}_j \right] \quad
\label{eq:HA}
\end{equation}
where $\mathbb{I}$ denotes the indicator function.
As each label is evaluated independently of the others, binary relevance methods typically perform quite well.

\emph{Subset accuracy} (SA), on the other hand, measures the ability of a classifiers to predict exactly the target label set.
\begin{equation}
    \text{SA}= \mathbb{I}\left[\vy = \hat{\vy}\right] \quad
\label{eq:SA}
\end{equation}
Thus, methods such as the label power set approach can be expected to perform comparatively well on this measure.
However, if the number of labels is rather large,
subset accuracy is often of limited use since most of the predictions will be wrong in at least one label, causing SA to evaluate to zero.

Hence, we additionally consider \emph{example-based F1} (F1)
to measure the performance.
Rooted in information retrieval, its key idea is to evaluate the harmonic mean between the precision (how many of the predicted labels are relevant?) and the recall (how many of the relevant labels have been predicted?) of the predicted labels for each example, averaged over all examples.
F1 can be considered as a compromise between HA and SA.
\begin{equation}
    \text{F1}=\frac{2 \sum_{j=1}^N y_j \hat{y}_j}{ \sum_{j=1}^N y_j + \sum_{j=1}^N \hat{y}_j} 
    \label{eq:F1}
\end{equation}

The measures also differ in the computational complexity they incur: in order to minimize SA, we must find the mode of the joint label distribution, whereas it is sufficient to find the modes of the marginal label distributions for HA. 
If there are dependencies between labels, both modes do not have to coincide.
\skp{In consequence,
CC\drin{, especially if using the same base learner and configuration as its BR counterpart,} cannot be expected to improve over BR regarding HA.
On the other hand, the reverse behaviour can be expected for SA. }
Hence, the trade-off between both measures and the relation to BR can serve to assess the ability of considering label dependencies.

As the objective of classifier chains is to find the correct label combination,
we expect the impact of our proposed extensions to be best reflected in SA. \raus{Conversely, we do not expect large differences between BR and CC in terms of HA, which evaluates all labels independently. }
The F1 measure is less extreme than SA, since it also considers partial matches and is therefore
often used
for providing a general comparison of the predictive quality.
Though it is sufficient to obtain good estimates for the individual labels in order to optimize univariate losses such as F-measure or Hamming loss \citep{dembczynski12PCCdependence}, this is not necessarily the case when there are dependencies between the labels.

\raus{
As \citet{dembczynski12PCCdependence} indicate, it is sufficient to obtain good estimates for the individual labels in order to optimize univariate losses such as F-measure or Hamming loss.
However, this is only valid if there are no dependencies between the labels.
However, as \citet{dembczynski12PCCdependence} indicate, it is sufficient to obtain good estimates for the individual labels in order to optimize univariate losses such as F-measure or Hamming loss.
Nevertheless, our approach may still benefit from the dependencies captured by the chaining approach with respect to these measure.
}

\subsection{Datasets}


\begin{table}[tbp]
\caption{Multilabel datasets used in this study, along with their \rev{respective} number of instances and labels, the average number of labels per instance (cardinality), and the number of distinct label combinations in the dataset.}
\label{tab:datasets}
\centering
\resizebox{\textwidth}{!}{
\begin{tabular}{lcccc|lcccc}
\toprule
Dataset	&	Instances	&	 Labels	&	Cardinality	&	Distinct 	&	Dataset	&	Instances	&	 Labels	&	Cardinality	&	Distinct \\
\midrule
\ds{emotions}	&	593	&	6	&	1.869	&	27	&	\ds{genbase}	&	662	&	27	&	1.252	&	32 \\
\ds{scene}	&	2407	&	6	&	1.074	&	15	&	\ds{medical}	&	978	&	45	&	1.245	&	94 \\
\ds{flags}	&	194	&	7	&	3.392	&	54	&	\ds{enron}	&	1702	&	53	&	3.378	&	753 \\
\ds{yeast}	&	2417	&	14	&	4.237	&	198	&	\ds{bibtex}	&	7395	&	159	&	2.402	&	2856\\
\ds{birds} 	&	645	&	19	&	1.014	&	133	&	\ds{CAL500} 	&	502	&	174	&	26.044	&	502 \\
\ds{tmc2007}	&	28596	&	22	&	2.158	&	1341	&		&		&		&		&	\\
\bottomrule
\end{tabular}
}
\end{table}

By now, there are many benchmark datasets for multi-label
classification available, which cover a wide variety of application areas.\footnote{Repositories of multi-label datasets can be found at \url{http://mulan.sf.net/datasets-mlc.html} and \url{http://www.uco.es/kdis/mllresources/}.}
From these, we selected the datasets shown in Table~\ref{tab:datasets}, which have various characteristics in terms of the number of instances and labels, the average cardinality of the relevant labels per example, and the number of distinct label sets that occur in the training data.
Not visible from the statistics (and generally not known) are correlations and dependencies between the labels.
As we have discussed above, chaining approaches can only be expected to gain an advantage over, e.g., BR if there are global or local dependencies between the labels in the dataset, which can be picked up and modeled by the learner.
For instance, there is evidence that \ds{yeast} and \ds{enron} contain mostly global dependencies whereas \ds{scene} also exhibits local dependencies \citep{papagiannopoulou15deterministicrelations,loza16MLRL,moyano17MLDA}.
Unfortunately, so far only few works have tried to systematically analyze  these characteristics.
All datasets came with predefined train-tests splits which were used for the evaluation.

\section{Dynamic Classifier Chains}
\label{sec:DCC}

Classifier chains depend on a fixed, static ordering of the labels. As we have discussed above (Section~\ref{sec:CC}), this problem is typically tackled by a heuristic choice of a suitable ordering, by pooling the result of multiple orderings, or by simply selecting a random ordering.
Apart from the computational disadvantages of exploring different label sequences,
the underlying assumption that there is one unique, globally optimal ordering which fits equally to all instances, can also be questioned. It is therefore natural to investigate whether a suitable classifier chain can be chosen
depending on the test instance at hand.

In this section, we will first motivate the need for dynamic classifier chains (Section~\ref{sec:DCC-Motivation}), then formalize the problem (Section~\ref{sec:DCC-Formal}), and finally review prior work in this area (Section~\ref{sec:DCC-Related}), before we introduce tree-based solutions to this problem in the following Sections~\ref{sec:dynamicpredictions} and~\ref{sec:DCC-XGBoost}.

\subsection{Motivation}
\label{sec:DCC-Motivation}

\begin{figure}[tbp]
    \centering
    \includegraphics[width=\textwidth]{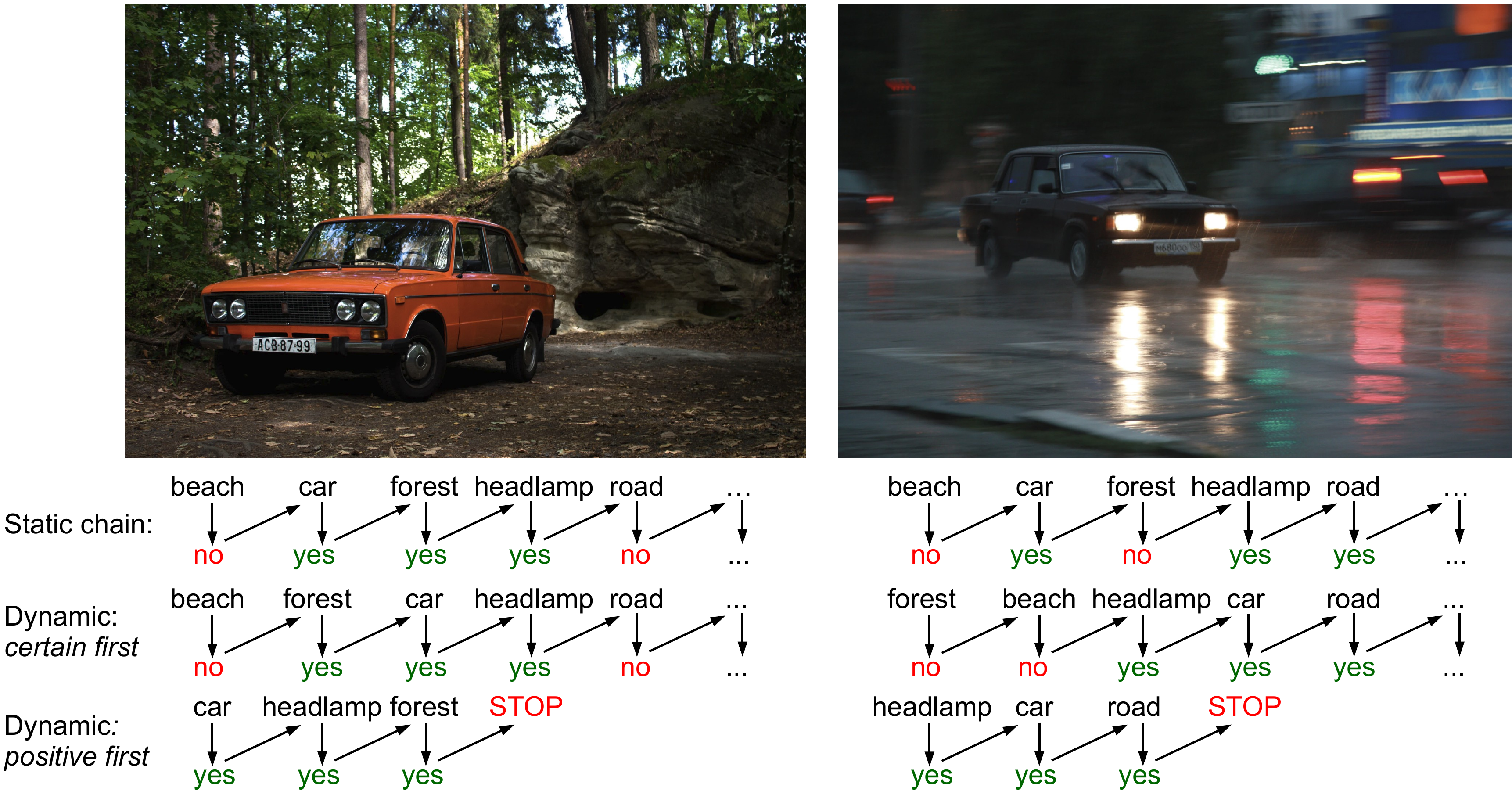}
    \caption{Example for different classifications using the static and the dynamic ordering for a two different pictures. See text for explanations.}
    \label{fig:illustration}
\end{figure}

Consider a hypothetical example of labeling an image with all objects that appear in it. More specifically, we want to identify cars and their parts in their surroundings. Figure~\ref{fig:illustration} shows two example images, which both show a car, one taken by day in a forest, and one by night in a city.
%
%
Classifier chains would predict the possible labels on both pictures in a static, heuristically or randomly selected order. For the sake of the example (and without loss of generality), we show an alphabetical ordering in the first row below the pictures. It is first predicted whether the scenery shows a beach or not, then whether it shows a car, a forest, a headlamp, a road, 
and so on.
Each prediction depends on the previous ones (and on the input image itself).
It seems natural to assume that a good order would predict the label for which it is most certain about first. Moreover, as a prediction for a label $\lambda_i$ influences the predictions of all subsequent labels $\lambda_j, j>i$, predicting the most certain labels first will also help to minimize the effects of error propagation along the chain.

However, clearly, if we try to sort the labels according to the certainty with which they can be recognized in each of the pictures, we obtain a different chain in each of the images. For example,
whereas predicting the presence of a \textsf{headlamp} can benefit from the previous rather simple detection of a \textsf{car} in the left picture, the opposite is the case in the dark picture on the right where the headlamps are considerably easier to detect than the car itself.
While the static chain of alphabetically ordered labels in the CC model depicted in the first row can only exploit the local dependency on the left picture, the dynamic approach illustrated in the second row,
which predicts first the labels for which it is most certain,
can also take advantage from the rather easy detection of \textsf{headlamp} on the right by leaving the prediction of \textsf{car} for later.

Note, however, that many of the certain predictions are actually negative. For example, we may quite reliably infer that the left pictures does not show a \textsf{beach}, or that the right picture does not show a \textsf{forest}. In many applications, such predictions are not desirable. So even though they might be helpful, it feels more natural
to have a chain that contains only positive dependencies.
Moreover, it is certainly more efficient, particularly considering that in MLC the number of positive labels usually stays in the tens even when the total number of labels is into the hundreds or thousands.
The approach in the last row chooses first the labels for which it is most certain that they are present.
After the predictions of \textsf{forest} and \textsf{road}, respectively, we can already stop the detection process since only negative predictions are to come.
The advantage comes at the expense of  ignoring dependencies to negative labels. Predicting the absence of a label is often much easier than finding positive ones and the knowledge about the absence of a label might be very useful to predict a positive label.
For instance, the detection of \textsf{forest} may benefit from the information that \textsf{beach} is not in the scenery of the left picture, as used by the second approach.

In the remainder of the paper, we will discuss tree-based approaches that are able to learn dynamic prediction chains, and, in one case, also focus on positive labels only. But first, we will formally define the problem.

\subsection{Problem Definition}
\label{sec:DCC-Formal}

From a formal point of view, adapting the order of the predicted labels simply corresponds to a context-dependent re-ordering or the chain rule.
More specifically, we can represent the joint distribution as
\begin{equation}
P(\vy \given \vx, \pi) =
\prod_{k=1}^N P(y_{\pi_k} \given  y_{\pi_k}, \ldots, y_{\pi_{k-1}}, \vx)
\end{equation}
where $\pi$ is a permutation over $N$ in one-line notation, i.e.,
$\{\pi_k \given 1 \leq k \leq N\}=\{1,\ldots,N\}$.

CC estimates the mode of $P(\vy \given \vx)$
by greedily maximizing $f_k(\vx,\hat{y}_{\pi_1},\ldots,\hat{y}_{\pi_{k-1}}) \approx P(y_{\pi_k} \given  y_{\pi_1}, \ldots, y_{\pi_{k-1}}, \vx)$ using a fixed, predetermined $\pi$.
A dynamic approach, instead, determines $\pi$ depending on the instance $\vx$ at hand.
For instance, labels could be ordered according to certainty, i.e., the closeness of the conditional probability to 0 or 1:
\begin{equation}\label{eq:DCCpicertainty}
\pi_k(\vx)=\argmax_{j\in \{1 \ldots N\}\backslash \{\pi_1(\vx) \ldots \pi_{k-1}(\vx)\}}  \left|0.5-P\left(y_j \given  y_{\pi_1(\vx)}, \ldots, y_{\pi_{k-1}(\vx)}, \vx\right)\right|
\end{equation}
as in the second setting depicted in  Figure~\ref{fig:illustration}.
Ordering according to descending probability, as for the model in the last row, corresponds to choosing $\pi$ as
\begin{equation}\label{eq:DCCpipositives}
\pi_k=\argmax_{j\in \{1 \ldots N\}\backslash \{\pi_1(\vx) \ldots \pi_{k-1}(\vx)\}}  P(y_j \given  y_{\pi_1(\vx)}, \ldots, y_{\pi_{k-1}(\vx)}, \vx)
\end{equation}
which would order the chain rule for an instance with $p$ positive labels as ($\pi(\vx)$ written as $\pi$ for convenience)
\begin{equation}\label{eq:DCCpositives}
P(\vy \given \vx, \pi) =
\prod_{k=1}^N P(y_{\pi_k} \given  \underbrace{y_{\pi_1}, \ldots, y_{\pi_{p}}}_{\text{positive labels}}, \underbrace{y_{\pi_{p+1}}, \ldots, y_{\pi_{k-1}}}_{\text{negative labels}}, \vx)
\end{equation}

Recall that theoretically, by the product chain rule, all decompositions are equivalent and independent of the chosen permutation \citep{cc_prob}, i.e.,
\begin{equation}
   P(\vy \given \vx, \pi) = P(\vy \given \vx, \pi')  = P(\vy \given \vx)
\end{equation}
for two different permutations $\pi$ and $\pi'$ of the label set.
However, in practice, the corresponding probability estimates and the resulting classifiers $f_k(\vx, \hat{y}_{\pi_1} \dots \hat{y}_{\pi_{k-1}})$ are error prone, in which case the order of the labelings does matter.
To see this, assume a simple problem where for each example either all labels are present ($\forall j: y_j = 1$), or all labels are absent ($\forall j: y_j = 0$). However, each of the labels is noisy, so that the BR classifiers $f_j(\vx)$ have different error rates $\epsilon_j$. Without loss of generality, let $\epsilon_1 < \epsilon_2 < \dots < \epsilon_N$. The HA~\eqref{eq:HA} of the binary relevance classifier is the average accuracy $\frac{1}{N}\sum_j(1 - \epsilon_j)$. Let us further assume that these errors are correlated, so that a classifier chain is able to pick up the signal that all subsequent classifiers just repeat the first label, i.e., $f_k(\vx, \hat{y}_{\pi_1} \dots \hat{y}_{\pi_{k-1}}) = \hat{y}_{\pi_1}$ for all $k > 1$. The accuracy of CC now clearly depends only on the error of the first member in the chain $\hat{y}_{\pi_1} = f_{\pi_{k-1}}(\vx)$, i.e., on $\epsilon_{\pi_1}$. If CC is trained using identity permutation ($\pi_k = k$), its HA will be $1-\epsilon_1$ and therefore higher than the one for BR. If CC is trained on the inverse permutation ($\pi_k = N+1-k$), its HA will be $1-\epsilon_N$, and therefore lower than BR.
The expected HA for a randomly selected chain order will be the same as the HA for BR.

So, in the above simple example, the performance of CC, and whether it is able to outperform BR, clearly depends on the chosen label ordering, and, coincidentally, the frequent default choices of selecting a random label order, or of averaging among several different orders, will, in expectation, not outperform BR.

If we now slightly modify the above example, so that label errors $\epsilon_j$ are not evenly distributed, but that certain labels can be better predicted in some regions of the example space, and \rev{worse} in other regions, we can, by essentially the same argumentation, arrive at the conclusion that an optimal classifier for the above problem needs to identify the most reliable label for a given test example, and start the chain with this label. This is essentially the idea of the approaches that we will discuss in the remainder of the paper.


\subsection{Related Work}
\label{sec:DCC-Related}

The idea that the optimal order in which labels are predicted may depend on the input example at hand is not entirely new, and has been tried in various settings before.
%
Da Silva et al. \citeyearpar{daSilva2014distinct}
made a first attempt by letting a nearest neighbor classifier decide which ordering to use for a given instance.
However, the dynamic selection was restricted to a pre-determined set of static label orderings.
It is also computationally expensive since new CC models have to be build during prediction.
%
\citet{jn:NIPS-17-MLC-RNN} use recurrent neural networks to predict the positive labels as a sequence,
but the ordering in which these are predicted is pre-determined.
%
\citet{jn:ICML-19} attempted to solve this problem using reinforcement learning and recurrent neural networks, but found that this does not outperform simple ordering strategies.
Moreover, the employed reinforcement learning approach essentially comes down again to exploring many possible label sequences during training, which is computationally very demanding.
\citet{llerena17multilabelSPN} propose several approaches for MLC based on \emph{sum-product networks}, including a sequential classification method which can either use a static or a dynamic ordering, which focuses on predicting positive labels first. However, the authors do not examine the difference between these two strategies. 
\citet{DCC-MLL} introduce dynamic chaining based on accuracy estimates of binary relevance predictions in a local neighborhood of the test example, and employ this technique for nearest neighbor and Na\"ive Bayes classifiers, but, in the latter case, have to make a conditional independence assumption on the label predictions as well.


\section{Dynamic Classifier Chains with Random Decision Trees}
\label{sec:dynamicpredictions}

In this section, we show how dynamic classifier chains can be efficiently constructed using
an extension of RDTs,
which
are  particularly appealing for this purpose:
First,  as we will see in the following, we can collapse BR and other MLC transformation or decomposition methods \citep{ml_tsoumakas_mining} such as CC to a single RDT ensemble without loss in predictive accuracy, therefore saving memory and computational costs. Second, and more importantly, RDTs can provide a controlled environment where we can compare alternative decomposition methods, prediction methods and other extensions isolated from any side effects since the model can be fixed beforehand and be the same for every analyzed approach.

\subsection{Random Decision Trees for Multi-label Classification}
\label{sec:RDT}

\citet{rdt_fan} introduced RDTs as an ensemble of randomly created decision trees, i.e., as trees for which the tests at the inner nodes are chosen randomly during the construction.
This is the major difference compared to classical decision tree algorithms, but also to the well known algorithm family of random forests~\citep{random_forest}, where only the subset of features which each tree learner can use is randomly drawn for each node.
In contrast, RDTs do not optimize any objective function during training, yet they are able to achieve competitive and robust performance \citep{rdt_hash}.
They were previously already successfully applied to MLC, focusing on large scale problems~\citep{rdt_ml} and streaming data with concept drift~\citep{rdt_stream}.

\subsubsection{Training}
\label{sec:training}

Starting from the root node, inner nodes of a random tree are constructed recursively by distributing the training instances according to the randomly chosen test at the inner node as long as the stopping criterion of maximum depth or minimum number of instances is not fulfilled. Discrete features are chosen without replacement for the tests in contrast to continuous features, for which additionally a randomly picked instance determines the threshold \citep{estimation_posterior}.
In case that no further tests can be created, a leaf will be constructed in which information about the assigned instances will be collected.
For MLC, we track the number of instances $N_{v}^\theta$ in leaf $v$ of tree $\theta$ in relation to the number of positive values $n_{v}^\theta(j)$ for label $\lambda_j$.

\subsubsection{Prediction}
\label{sec:RDTprediction}

During prediction, an instance is forwarded from the root to a leaf node passing the respective tests in the inner nodes. In case some of the tested features have missing values, all branches are visited and the function
$
q(\vx,\theta) \subset \{1,\ldots,|\theta|\}$
returns a set
of the leaf indices in tree $\theta$ to which the instance has been assigned to. Following \citet{estimation_posterior}, the posterior probability that the specific label $y_{j}$ is true given an instance $\vx$ and a tree $\theta$ can be
formalized as
\begin{equation}\label{eq:probestimates}
P(y_{j}=1 | \vx, \theta) = \frac{ \sum_{v \in  q(\theta, \vx)} n_{v}^\theta(j)}{\sum_{v \in  q(\vx,\theta)} N_{v}^\theta}
\end{equation}


As the randomness results in a large variety of distribution in an ensemble of RDT, many of them approaching the prior label distribution, we propose to distinguish between the quality of the collected statistics and to reward trees with higher confidences in their estimates.\com{vlt. hier eine ref zum DS paper, wo es ja ein wenig genauer analysiert wurde} The \emph{Gini index} is often used for determining the purity of a distribution, which we use in inverted form as follows
\begin{equation}
	w(\vx, \theta) = 1 - \frac{4}{N} \sum_{j=1}^N P(y_{j}=1 | \vx, \theta) \left(1- P(y_{j}=1 | \vx, \theta)\right)
\end{equation}
in order to weight the estimates of the individual trees, resulting in the overall prediction for the ensemble $\Theta$ as
\begin{equation}
\label{eq:weightedensembleestimates}
f_j(\vx) = P(y_{j}=1 | \vx, \Theta) = \frac{1}{\sum_{\theta \in  \Theta} w(\vx, \theta) }\sum_{\theta \in  \Theta} P(y_{j}=1 | \vx, \theta) w(\vx, \theta)
\end{equation}

An obvious option in order to obtain multi-label predictions from the estimations in \eqref{eq:weightedensembleestimates} is to use a threshold of $50\%$ so that $\hat y_j = \mathbb{I}\left[f_j(\vx) \geq 0.5\right]$.
However, as \citet{thresholding_strategy} observed, a threshold of $50$\% is not always ideal. Note that the tests in the tree are not specifically chosen to obtain a high purity of the distributions in the leaves, and in fact many leaves might contribute only with estimates close to the prior distribution, pulling down the average estimates. Thus, we follow \citet{rdt_ml} and estimate the average number of relevant labels as
\begin{equation}
R(\vx,\Theta) = \frac{1}{\sum_{\theta \in  \Theta} w(\vx, \theta) } \sum_{\theta \in  \Theta} r(\vx,\theta) w(\vx, \theta)
\end{equation}
where
\begin{equation}
\label{label_method_1}
r(\vx,\theta) = \frac{ \sum_{v \in  q(\vx, \theta)} \sum_{j=1}^n  n_{v}^\theta(j)}{\sum_{v \in  q(\vx,\theta)} N_{v}^\theta}.
\end{equation}
$R(\vx,\Theta)$ is rounded in order to get an integer. This value is then used to cut the ranking of labels induced by the distribution of the marginals $P(y_j=1 \given \vx, \Theta)$. 

\subsection{Static Chain Ordering\com{or: Integrating the chain}}
\label{sec:collapsedCC}
The use of RDTs allows us to collapse a classifier chain to a single RDT in the following way:
Instead of training $N$ RDTs on the $N$ augmented spaces
$X^{(k)} = [X,Y_{\cdot,\pi_1},\ldots, Y_{\cdot,\pi_{k-1}}]$,
we train only one RDT on the complete augmented space $[X,Y] \in \mathcal{X} \times \mathcal{Y}$.
This type of RDT can answer any query in the form of
\begin{equation}
P(y_{\pi_k} \given y_{\pi_1}=b_1, \ldots, y_{\pi_{k-1}}=b_{k-1},\vx),\ b_{j} \in \{0,1\}
\end{equation}
by creating a query instance $(\vx,\vp)$ where the $p_{\pi_1}=b_1, \dots, p_{\pi_{k-1}}=b_{k-1}$ are filled with the available label values, and all remaining $p_j$ 
set as unknown or missing.
The RDT can then answer the query by combining all possible paths for the missing values, as described in Section~\ref{sec:RDTprediction}.
A classifier chain prediction for a fixed label ordering $\pi$ is hence obtained by initializing $\vp^0$ with
missing values
and filling up $p^k_{\pi_k}$
with the label predictions $\hat{y}_{\pi_k}=\mathbb{I}\left[f_{\pi_k}(\vx,\vp^{k-1})\right] \geq 0.5]$ while proceeding through the chain.

As RDTs are completely randomized, we can
expect on average the same predictions
as for a RDT ensemble which skipped the respective feature during training.
In fact, in our experiments, we control  the percentage of activated label tests with a parameter $\sigma$, which allows us to analyze the effect of using previous predictions on an otherwise unchanged model.

\begin{figure}[!tb]
	\center
	\includegraphics[width=0.6\textwidth]{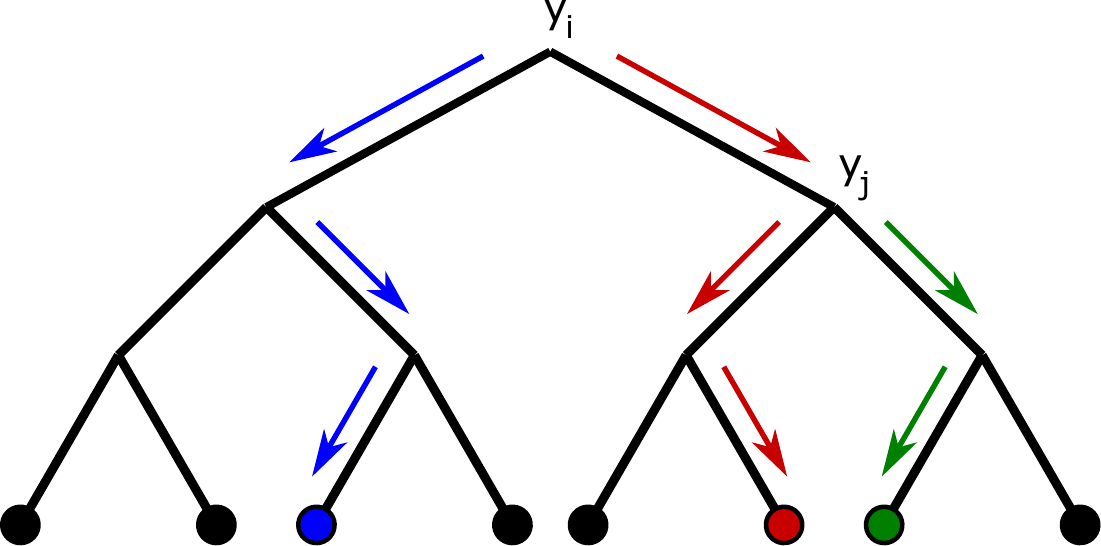}
	\caption{Example for the refinement of a prediction for a particular instance and decision tree.
$y_i$ and $y_j$ indicate tests on labels at the respective inner nodes.}
	\label{pic_example_prediction}
\end{figure}

Figure~\ref{pic_example_prediction} visualizes the prediction process for a label on a single tree: Let us assume that the label to be predicted is $y_i$, which comes before $y_j$. In this case neither $y_i$ nor $y_j$ are known, i.e. all three colored branches are followed and the respective leaves are used in order to produce a prediction for $y_i$. For label $y_j$ the previous label $y_i$ would be known, so that---depending on its value---we would skip either the left or the right branch, obtaining a label distribution at the leaves which is different and more refined than the previous one. Indeed, we can observe that the number of leaves on which the prediction relies, is monotonically decreasing during the classification process. Therefore, the set of leaves to which the instance is assigned in the first iteration will always be a superset of the leaves of the following iterations. This leads to a refinement of the predictions throughout the iterations.

\subsection{Dynamic Chain Ordering}
\label{sec:DCC-ordering}

In order to take advantage of the situation that predicting a label before or after another one might be easier depending on the instance at hand, we propose to let the RDT decide which label to predict next. Hence, instead of using the
estimated probabilities
to decide whether the $i$-th label in $\pi$ is positive or negative, we use it to set the label for which RDT is most confident in its prediction. Labels, which were already predicted, are ignored.

Starting with the empty prediction vector $\vp^0$,
in each iteration $i$ we select the label for which the RDT is most confident following \eqref{eq:DCCpicertainty}:
\begin{equation}
		\pi_k = \argmax_{j \in \{1, \ldots, N\}\backslash \{\pi_1, \ldots, \pi_{k-1}\}} \left|0.5 - f_{j}(\vx,\vp^{k-1})\right|
\end{equation}

Let us consider again the tree in Figure~\ref{pic_example_prediction}. The difference to the static chain approach is that the aggregated blue, red and green leaves would be used in order to determine whatever label $y_k$ is most likely given the found distribution, instead of a specific label (in the previous example label $y_i$). Hence, the RDT could decide to predict $y_j$ instead if they are more confident about it, or any other label with the highest confidence.
\skp{We believe that this potentially leads to more reliable predictions, both in terms of individual labels as well as label combinations.}


The process of predicting the value needs further adaption due to the iterative prediction of the labels.
The idea is to have predicted exactly $R((\vx,\vp^N),\Theta)$  \rev{positive labels} after the prediction sequence is completed.
Since the prediction $\vp$ is constantly changed during the classification process, $R((\vx,\vp^k),\Theta)$ has to be re-computed in every iteration.
First of all, we can only predict a label in iteration $k$ positive if the number of already predicted positive labels $|\vp^{k-1}|$ is smaller than $R((\vx,\vp^{k-1}),\Theta)$. Moreover, we have to predict a label as positive if we know that all the remaining labels in the chain need to be predicted positive to ensure that we obtain exactly $R((\vx,\vp^{k-1}),\Theta)$ positive labels.
Hence, the $k$-th label is predicted as
\begin{equation}
\label{label_method_4}
p_{\pi_k} =
\begin{cases}
    1,& \text{if } P\left(y_{\pi_k}=1 \given (\vx,\vp^{k-1}),\Theta\right)\geq 
    0.5 \text{ and } \left|\vp^{k-1}\right|  < R\left((\vx,\vp^{k-1}),\Theta\right) \\
		1,& \text{if } N-k < R\left((\vx,\vp^{k-1}),\Theta\right) - \left|\vp^{k-1}\right|  \\
    0,& \text{otherwise}
\end{cases}
\end{equation}

\subsection{Evaluation}
\label{sec:RDTevaluation}

A key aspect in our experimental evaluation was to verify our ideas of dynamic classifier chains on the usage of RDT as a controlled experimental environment for fair and specific comparisons. In particular, the focus was to demonstrate that using dynamic, context-dependent predictions improves over using static orderings w.r.t. predictive performance (Section \ref{sec:rdt_CCvsDCC}). A decisive role in this is played by the influence of the previous predictions on the current prediction, which is analyzed in Section~\ref{sec:rdt_influence_label_tests}. Lastly in Section~\ref{sec:rdt_dynamic_sequence}, we inspect the dynamic sequences in detail by evaluating the predicted labels in each iteration.



Unless otherwise noted, we have chosen to evaluate the parameter setting
of using 300 decision trees, a maximum depth of $30$, 
a maximum leaf size of $5$ and a percentage of label tests of $30\%$. Preliminary experiments with RDT revealed reasonable and stable performance for this parameter setting also on other kind of problems. 

%

\subsubsection{Static vs. Dynamic Label Orderings}
\label{sec:rdt_CCvsDCC}

\begin{table}
    \centering
    \caption{Comparison between the dynamic and  the static chain method.}
    \label{tab:rdt_static_dynamic}
    \begin{tabular}{l|cc|cc|cc}
\toprule
& \multicolumn{2}{c|}{HA} & \multicolumn{2}{c|}{SA} & \multicolumn{2}{c}{F1} \\
 &     static &    dynamic &     static &    dynamic &     static &    dynamic \\
\midrule
\ds{flags}    & 0.7299 & 0.7582 & 0.1292 & 0.1846 & 0.7126 & 0.7478 \\
\ds{emotions} & 0.6338 & 0.7632 & 0.0772 & 0.2525 & 0.3740 & 0.6228 \\
\ds{scene}    & 0.7174 & 0.8917 & 0.1594 & 0.6421 & 0.1832 & 0.6929 \\
\ds{yeast}    & 0.6907 & 0.7837 & 0.0146 & 0.2039 & 0.4617 & 0.6270 \\
\midrule
\ds{birds}    & 0.9205 & 0.9451 & 0.3140 & 0.4520 & 0.3505 & 0.5706 \\
\ds{cal500}   & 0.7933 & 0.8435 & 0.0000 & 0.0000 & 0.2946 & 0.4660 \\
\ds{enron}    & 0.9170 & 0.9374 & 0.0603 & 0.0656 & 0.2925 & 0.4038 \\
\ds{medical}  & 0.9504 & 0.9618 & 0.0033 & 0.1550 & 0.0036 & 0.2235 \\
\midrule
\ds{genbase}  & 0.9279 & 0.9375 & 0.1372 & 0.2714 & 0.1401 & 0.2714 \\
\ds{bibtex}   & 0.9725 & 0.9755 & 0.0000 & 0.0000 & 0.0136 & 0.1330 \\
\ds{tmc2007}  & 0.8691 & 0.8854 & 0.0053 & 0.0353 & 0.2486 & 0.3436 \\
\midrule
win/draw/loss & 0/0/11  & 11/0/0  & 0/2/9  & 9/2/0  & 0/0/11  & 11/0/0\\
\bottomrule
\end{tabular}
\end{table}

In this experiment we evaluated the advantage of the dynamic chain ordering in comparison to using a static chain ordering. Taking advantage of our controlled environment, we built for both approaches the same ensemble of trees, respectively. The only difference between the dynamic and the static setup is the ordering of the labels during the prediction process. We compare our proposed dynamic method to the averages over ten randomly-drawn but fixed orderings used for the static CC approach in Table~\ref{tab:rdt_static_dynamic}.

The first and foremost observation is that the dynamic chain ordering is clearly superior to the static chain ordering on all datasets.
This confirms our main hypothesis that it is advantageous to adapt the prediction order according to the context at hand.
In fact,
on most datasets the results for SA and F1 often doubles by using the dynamic chain instead of the static orderings.
However,
with respect to HA we can observe that only minor improvements can be achieved on the sparse datasets, that is,  \ds{enron}, \ds{medical}, \ds{genbase}, \ds{bibtex} and \ds{tmc2007}.

\begin{figure}[tb]
    \centering
    \includegraphics[width=\textwidth]{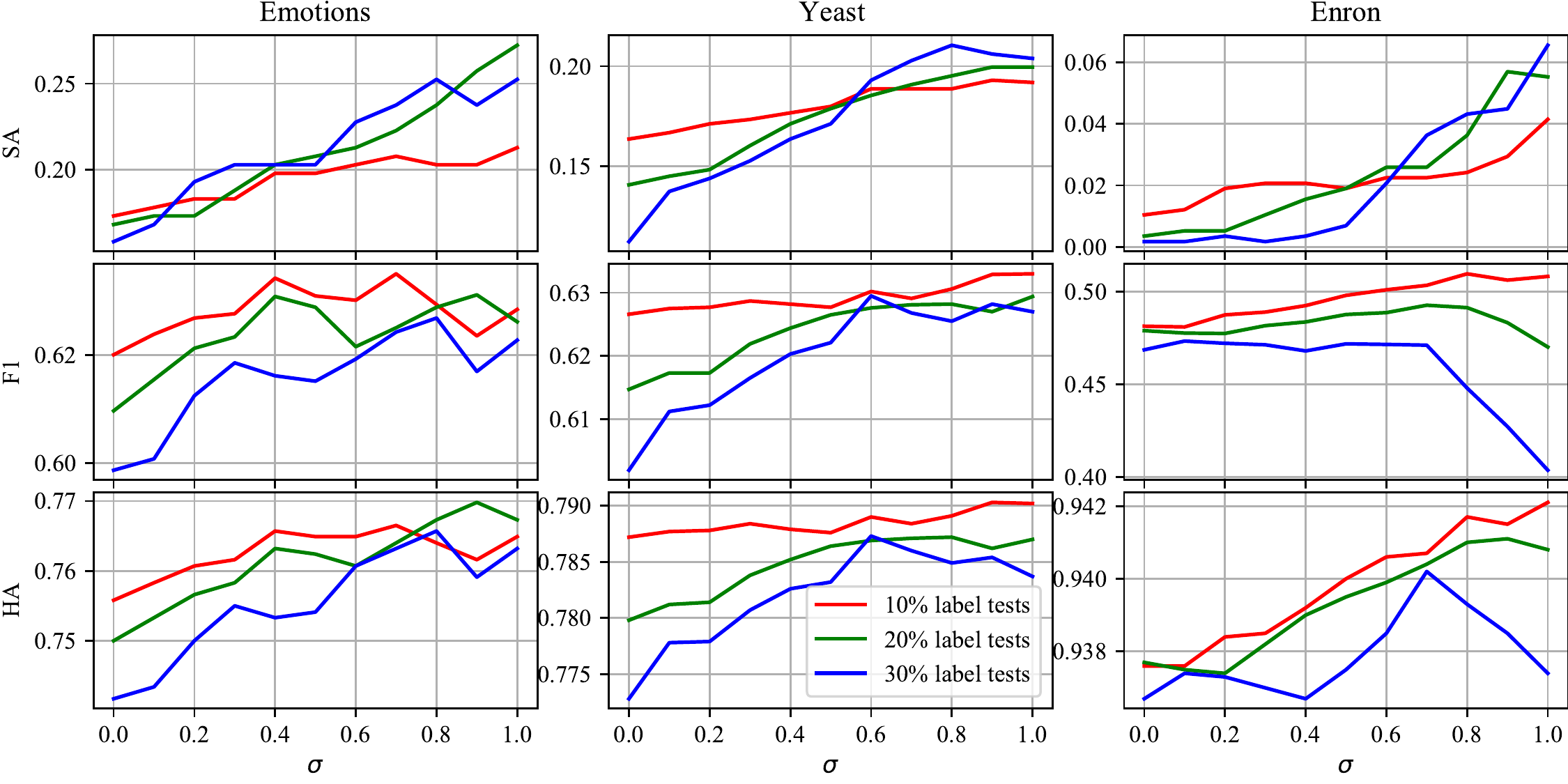}
    \caption{Influence of label tests on DCC. The $y$-axis represents the value for the measure and the $x$-axis represents the percentage $\sigma$ of activated label tests. The color indicates the percentage of label tests per tree.}
    \label{fig:rdt_exploiting_predictions}
\end{figure}


\subsubsection{Independent Predictions vs. Exploiting Previous Predictions}
\label{sec:rdt_influence_label_tests}

In this experiment we evaluated how the prediction is influenced by the usage of the label tests, i.e., by the usage of the previous predictions in the dynamic chain. At this stage the flexibility of the RDT algorithm pays off since we can choose the ratio $\sigma$ of activated tests on the labels without the need for adaptations of the model (cf. Section~\ref{sec:collapsedCC}). Hence, $\sigma=0$ corresponds to a binary relevance classifier using RDT
(more specifically, the collapsed version).
Incrementing $\sigma$ allows to directly observe utility and the effectiveness of exploiting potential label dependencies.
Furthermore, we directly control the probability of choosing a test on a label feature at the inner nodes (10\%, 20\% and 30\%).

Figure~\ref{fig:rdt_exploiting_predictions} shows the benefit for some selected datasets w.r.t. all evaluation measures (visualizations for all other datasets can be found in Figure~\ref{fig:rdt_influence_label_tests_appendix} in the appendix). For instance, we can observe on datasets \ds{emotions} and \ds{yeast} a major influence of the activated label tests on the performance.
It seems obvious
that there is a strong dependency between the labels in the datasets of which we can take advantage.
The positive effect is less pronounced on some of the other datasets, depending on the measure and the label test configuration.
For instance, we can observe a decrease in F1 for growing number of activated tests for \ds{enron} especially for the 30\% label test configuration.
However, the ability of predicting the correct label combination (SA) does not seem to suffer.

\subsubsection{Analysis of the Dynamic Sequences}
\label{sec:rdt_dynamic_sequence}

Our approach dynamically produces a different prediction sequence on the labels for each given test instance. We were interested in characterizing and analyzing these sequences, which were selected by the RDT as being most appropriate for producing accurate predictions.

\begin{figure}[!tb]
        \centering
        \begin{subfigure}[!ht]{0.5\textwidth}
                \includegraphics[width=\textwidth]{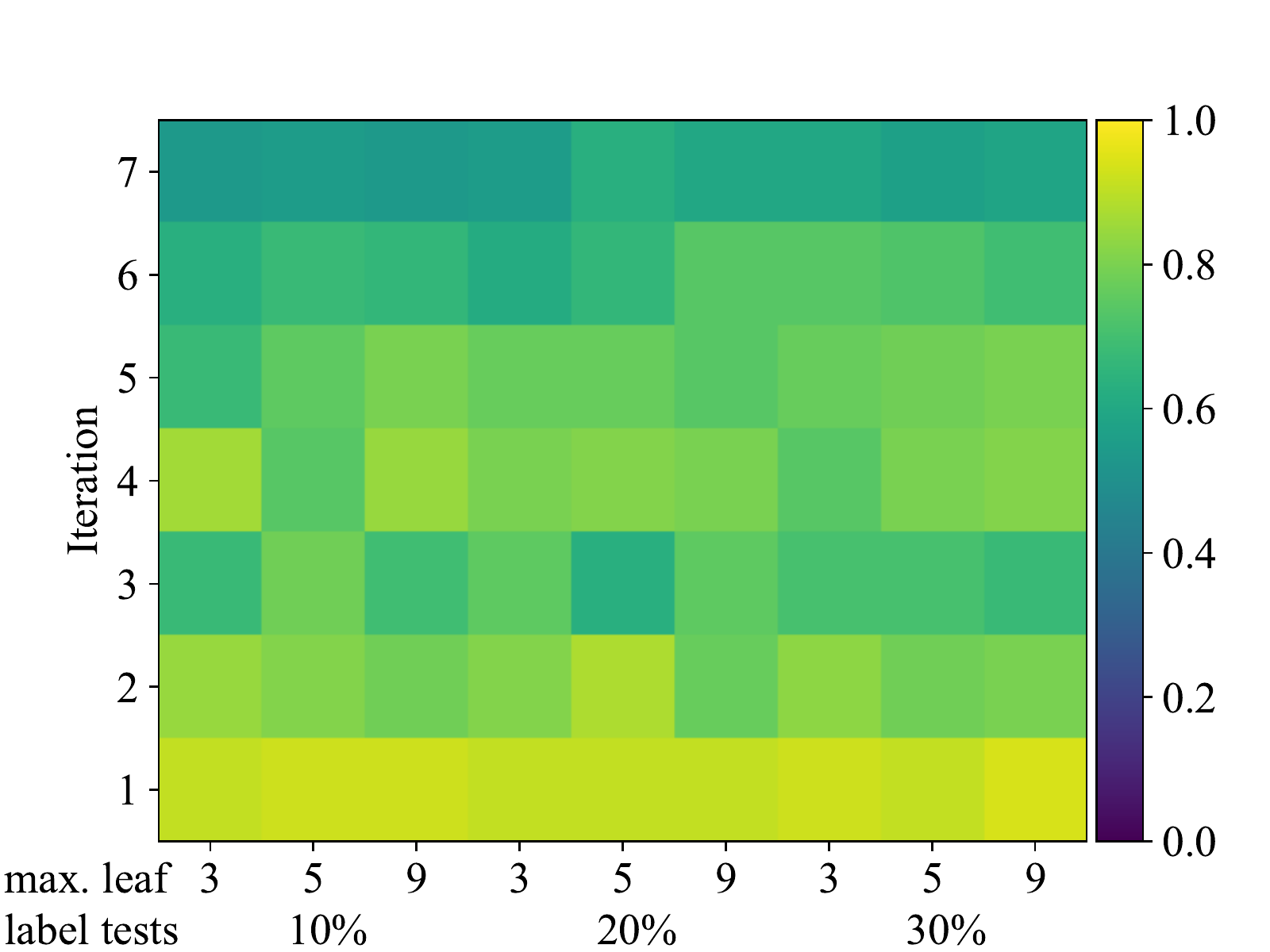}
                \caption{Accuracy of the prediction.}
                \label{fig:rdt_dynamic_squence_flags1}
        \end{subfigure}%
        \hfill
        \begin{subfigure}[!ht]{0.5\textwidth}
                \includegraphics[width=\textwidth]{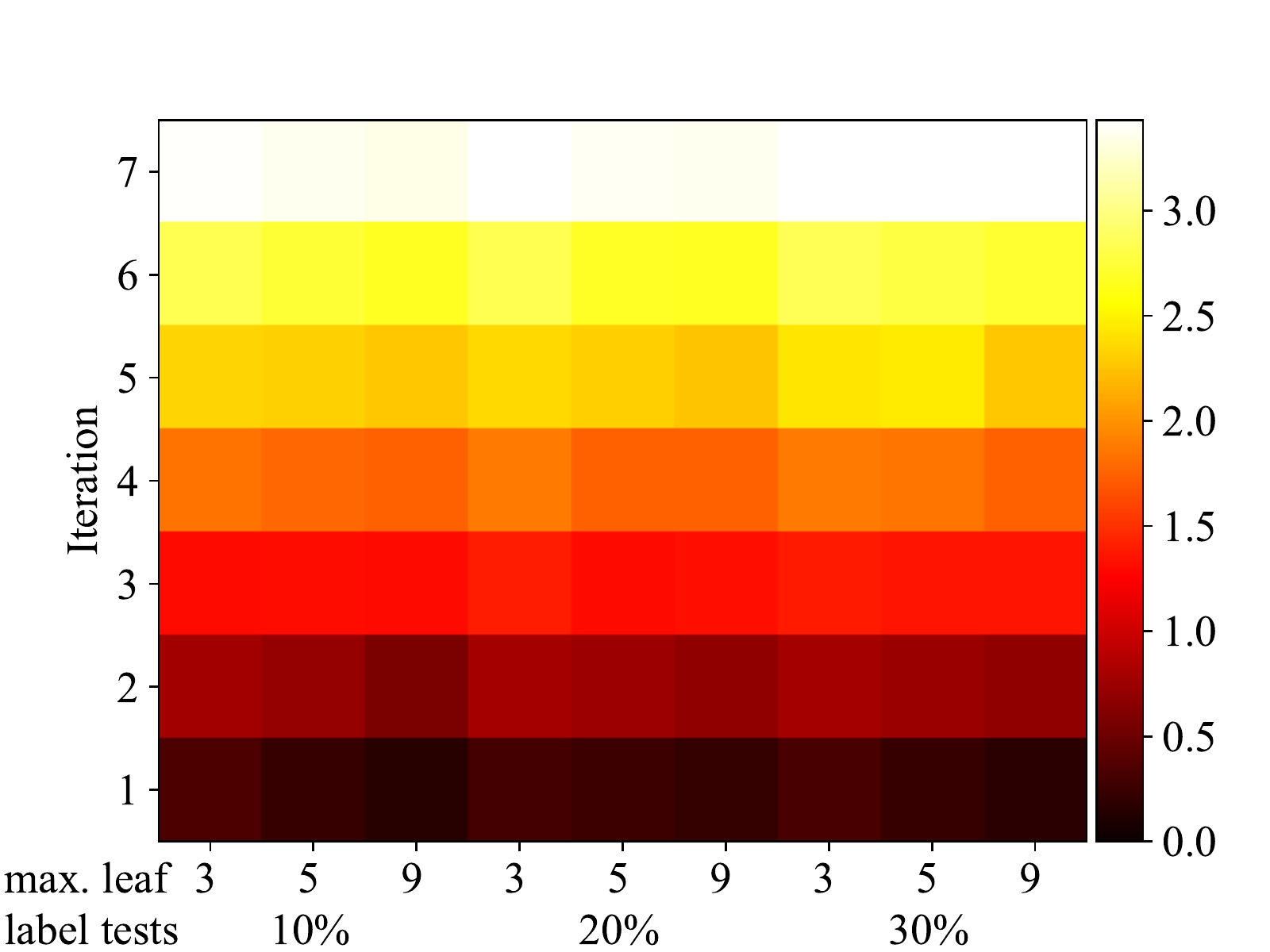}
                \caption{Number of positive labels.}
                \label{fig:rdt_dynamic_squence_flags2}
        \end{subfigure}
        \caption{Heatmaps characterizing the predicted sequences on \ds{flags}.}
        \label{fig:rdt_dynamic_squence_flags}
\end{figure}

\begin{figure}[!tb]
        \centering
        \begin{subfigure}[!ht]{0.5\textwidth}
                \includegraphics[width=\textwidth]{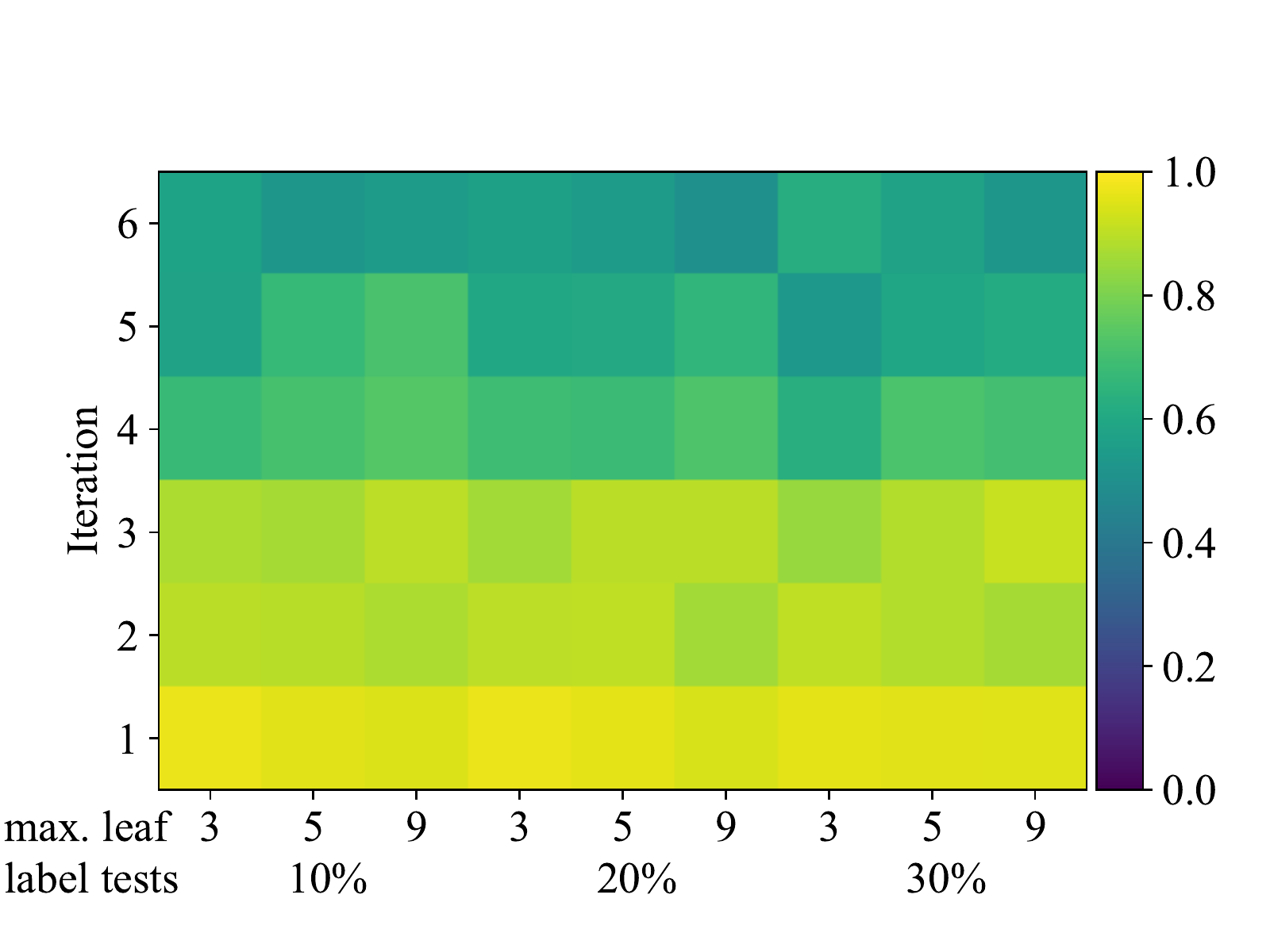}
                \caption{Accuracy of the prediction.}
                \label{fig:rdt_dynamic_squence_emotions1}
        \end{subfigure}%
        \hfill
        \begin{subfigure}[!ht]{0.5\textwidth}
                \includegraphics[width=\textwidth]{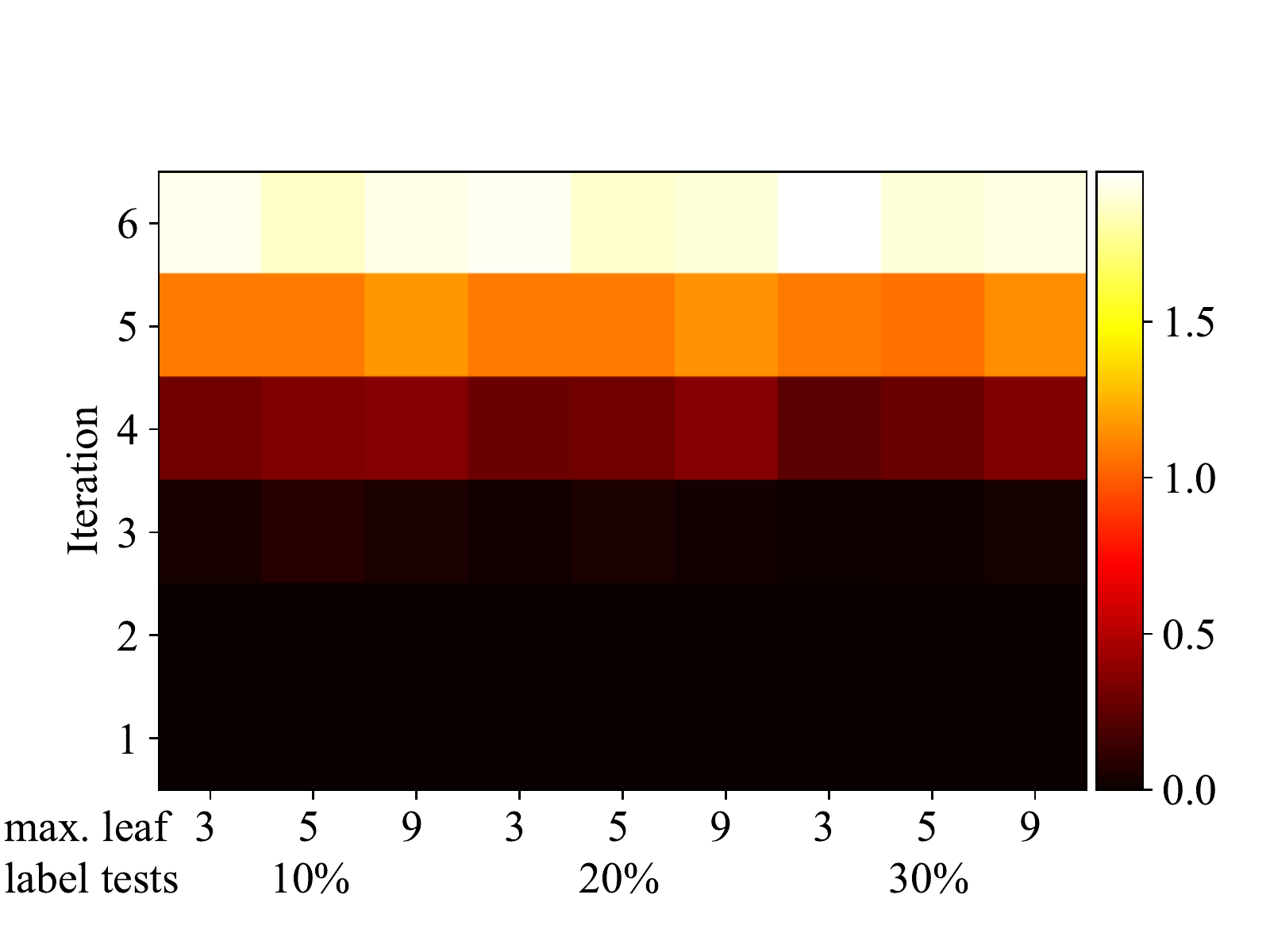}
                \caption{Number of positive labels.}
                \label{fig:rdt_dynamic_squence_emotions2}
        \end{subfigure}
        \caption{Heatmaps characterizing the predicted sequences on \ds{emotions}.}
        \label{fig:rdt_dynamic_squence_emotions}
\end{figure}

Figures~\ref{fig:rdt_dynamic_squence_flags} and \ref{fig:rdt_dynamic_squence_emotions} visualize our results exemplarily for \ds{flags} and \ds{emotions}. The heat map on the left shows the average accuracy (color) of predicting the $j$-th label in the dynamic sequence ($y$-axis) for different parameter configurations ($x$-axis), whereas the right map visualizes the number of labels (color) which were predicted as positive until a certain iteration.

We can observe on \ds{flags} and \ds{emotions}, as well as on the remaining datasets, that independently of the parameter configuration the predictions of the first iterations are much more accurate than the predictions at the end.
One reason for this picture is, of course, that our label selection method specifically chooses the labels where the RDT ensemble is most confident first.
This is also reflected by the heat maps on the right.
They show that RDTs tend to first predict the (easier) negative labels before heading to the (more difficult) positives ones.
Apparently, collecting as much as possible of the more readily accessible evidence  helps to take the harder decisions later in the chain, as the comparison to the static chains in Section~\ref{sec:rdt_CCvsDCC} demonstrates.
This is the case even though later predictions may suffer more from error propagation.

\subsection{Discussion}

Our experimental results based on the controlled evaluation environment of RDTs show that dynamic orderings improve over static orderings. In particular, the dynamic approach takes advantage by first predicting labels for which it is the most certain, which are then used to refine the estimates of the remaining, more challenging, labels in the following iterations.

With respect to computational complexity, the costs for building the trees and performing the dynamic predictions is
essentially the same as for
a static ordering. They mainly depend on the size of the ensemble and the depth of the trees. Moreover, the dynamic approach potentially allows to shorten the prediction process, namely when enough positive (or negative) labels have been already predicted,  removing the dependencies on the label size.

However, as preliminary experiments with RDT have shown this strategy was consistently inferior to selecting according to the certainty.
The reason might be again the non-optimized construction of the trees  which leads to underestimated probabilities close to the prior that are difficult to differentiate and select from.
An additional disadvantage due to the lack of any optimization is the potential performance gap to state-of-the-art methods.
For instance, our results so far and additional comparisons to state-of-the-art approaches (cf. Section~\ref{sec:comparisonbaselines}) show that RDTs are inherently not suitable for sparse data like text.
To address this problem, we propose in the next section to replace the simple and flexible training process of RDTs by a tree induction method which specifically optimizes the dynamic chain prediction.

\section{Learning a Dynamic Chain of Boosted Tree Classifiers \com{6p max}}
\label{sec:DCC-XGBoost}
As shown in the previous section, choosing the order of the labels dependent on the instance at hand can lead to a significant improvement in predictive accuracy, and dynamic classifier chains are able to exploit this.
An additional potential advantage of DCC originates from the fact that multi-label problems may include a large number of labels, but the actual number of assigned labels to instances almost always stays low.
As argued in Section~\ref{sec:DCC}, in such cases focusing on the positive labels could lead to a massive reduction in computational costs.
For instance, in a MLC dataset with 100 labels but maximum number of assigned labels of 5, restricting the length of the chain to this number could be sufficient for the classification problem whereas a CC would still have to train 100 models and perform 100 classifications per instance.
However, the advantage comes at the expense of not being able to consider dependencies to negative labels.
In order to be still able to exploit the computational advantage, we need more reliable and targeted predictions for the positive labels than the ones that RDT could provide in our experimental results.

The highly optimized gradient boosting framework \citep{friedman02gradientboosting} is well suited for this task since it allows to maximize arbitrary differentiable objective functions. 
This is achieved by adding decision trees  to an ensemble which are optimized in a gradient descent like procedure.
Furthermore, its iterative procedure seems adequate in order to naturally integrate the dynamic chaining process.
Similarly to DCC,
gradient boosted trees
are constructed by iteratively adding
trees which refine previous predictions in a step-wise manner. 
The proposed XDCC approach integrates DCC into the state-of-the-art extreme boosting architecture of XGBoost (Section~\ref{sec:XDCC}).
As a preparatory step, we propose the extension ML-XGB of XGBoost which is able to perform multi-label instead of only binary predictions at the leaves (Section~\ref{sec:ML-XGB}).

	\raus{Instead of learning a static CC that predicts labels in a predefined rigid order, we introduce a dynamic classifier chain (DCC) where each chain-classifier predicts only a single label which is not predetermined and can be different for each sample.
	To prevent learning bad label dependencies\tod{verstehe ich nicht, würe ich rauslassen}, the base-classifiers are built to maximize the probability of only a single label. On the one hand this allows to exploit more complex label dependencies,\tod{bis hier würde ich weglassen, und folgendes argument verwenden} and on the other side to massively reduce the length of the chain, while still being able to predict all labels. Given a dataset with 100 labels and a cardinality of five, a DCC of length five has the ability to predict all labels, whereas a CC would have to train 100 models.
	Because of XGBoosts highly optimized boosting-tree architecture, we decided to use it as base-classifiers in our chain. Therefore we have to modify it and make it capable of building multi-label-trees which can deal with an arbitrary number of labels.
}

\subsection{Extreme Gradient Boosted Trees
}
Extreme Gradient Boosted Trees 
\citep[XGBoost;][]{chen2016xgboost}
is a versatile implementation of gradient boosted trees. 
One of the reasons for its success is the very good scalability due to the specific usage of advanced techniques for dealing with large scale data.
XGBoost was originally designed for dealing with regression problems, but different objectives can be defined by correspondingly adapting the objective function and the interpretation of the numeric estimates.
Each model consists of a predefined number of decision trees. These trees are built using gradient boosting, i.e., the model is step-wise adding trees which further minimize the training loss.
The main difference to RDT tree construction is the way the feature splits inside the nodes are determined. RDT uses completely random split tests, whereas XGBoost aims for finding splits that maximizes a gain score for the resulting leaves.
The trees are constructed recursively, starting at the root node, by adding feature tests on the inner nodes.
At each inner node, all possible feature tests are evaluated according to the gain obtained by applying the split on the data.
The test candidate returning the highest gain score is then taken and both children are further split up until the maximum depth is reached or the gains stay below a certain threshold. A prediction can be calculated by passing an instance through all trees and summing up their respective leaf scores.

\paragraph{Boosted Optimization}
We refer to
\citep{chen2016xgboost}
for a more detailed description of XGBoost.
An XGBoost model consists of a sequence
$\theta^{(1)}, \ldots, \theta^{(T)}, T=|\Theta|$
of decision trees. 
Each tree $\theta^t$ returns a numeric estimate $f^{(t)}(\vx)$ for a given instance $\vx$.
Predictions are generated by passing an instance through all trees and summing up their leaf scores. The model is trained in an additive manner and each boosting round adds a new tree that improves the model most. For the $t$-th tree the loss to minimize becomes
\begin{align}\label{eq:XGBlossmin}
	L^{(t)} = \sum_{i=1}^M l\left(y_i,\ (\hat{y}_i^{(t-1)}+f^{(t)}(\vx_i))\right)+\Omega(f^{(t)}),
\end{align}
where $\hat{y}_i^{(t-1)}=\sum_{t'=1}^{t-1} f^{(t')}(\vx_i)$ is the prediction of the tree ensemble so far, $l(y,\hat{y})$ is the loss function for each individual prediction  and  $\Omega$ is an additional term to regularize the tree.
Combined with a convex differential loss function the objective can be simplified by a second-order approximation. This comes down to the following objective which can be minimized recursively at each node 
\begin{align}
		\label{eq:obj}
		{obj}^* &= -\frac{1}{2} \sum_{v=1}^{|\theta|} \frac{G_v^2}{H_v + \epsilon} + \gamma |\theta| \hspace{1cm}\text{with} \quad &G_v = \sum_{i \in I_v} g_i \text{, } H_v = \sum_{i \in I_v} h_i 
\end{align}
where $|\theta|$ is the size of the tree in number of leafs, $G_v$ defines the sum of the gradients for all instances $I_v$ in leaf $v$, $H_v$ is the corresponding sum of the Hessians (cf. also Section~\ref{sec:ML-XGB}),
and $\epsilon$ and $\gamma$ are regularization terms derived from $\Omega$.
In such a leaf, the predicted score $w_v^* = -\frac{G_v}{H_v + \epsilon}$ minimizes ${obj}^*$.
The following gain function is used to evaluate different splits at inner nodes, where $u$ and $v$ for $G$ and $H$ refer to the resulting left and right leafs.
\begin{align}
	L_{split} &= \frac{1}{2}\left[\frac{G_u^2}{H_u + \epsilon} +  \frac{G_v^2}{H_v + \epsilon} - \frac{(G_u + G_v)^2}{H_u + H_v + \epsilon}\right] -\gamma.
\label{eq:gain}
\end{align}

\raus{Combined with a convex differential loss function the objective can be simplified by taking the second-order approximation which gives us the final objective to optimize:
\begin{align*}		
	{obj}^{(t)} &= \sum_{v=1}^T[G_v w_v + \frac{1}{2}(H_v +\epsilon)w_v^2] + \gamma T
    \quad \text{with} &G_v = \sum_{i \in I_v} g_i \text{, } H_v = \sum_{i \in I_v} h_i ,
\end{align*}
where  $I_v$
is the set of indices for all data points in leaf $v$, $G_v$ defines the sum of the gradients for all instances $I_v$ in leaf $v$, $H_v$ is the corresponding sum of the Hessians (cf. also Section~\ref{sec:ML-XGB}), $w_v$ is the vector of leaf scores and $\epsilon$ and $\gamma$ are regularization terms derived from $\Omega$. With the optimal weights $w_v^*$ for leaf $v$ the objective becomes
\begin{align}
		{obj}^* &= -\frac{1}{2} \sum_{v=1}^T \frac{G_v^2}{H_v + \epsilon} + \gamma T \hspace{1cm}\text{with}\hspace{0.7cm} w_v^* = -\frac{G_v}{H_v + \epsilon}.
\end{align}
These weights finally lead to the gain function used to evaluate different splits. The indices $L$ and $R$ for $G$ and $H$ refer to the proposed right and left child candidates:
\begin{align}
	L_{split} &= \frac{1}{2}\left[\frac{G_L^2}{H_L + \epsilon} +  \frac{G_R^2}{H_R + \epsilon} - \frac{(G_L + G_R)^2}{H_L + H_R + \epsilon}\right] -\gamma.
\end{align}
}

	\subsection{Multi-label XGBoost}
	\label{sec:ML-XGB}
Since XGBoost only supports binary classification with its trees in the original implementation, the underlying tree structure had to be adapted in order to support multi-label targets.

The first modification is to calculate leaf weights and gradients over all class labels instead of only a single one.
More specifically, $G_{j,v} = \sum_{i \in I_v} g_{j,i}$ and $H_{j,v} = \sum_{i \in I_v} h_{j,i}$ extend to the labels $1\leq j \leq N$, abbreviated as $G_{j}$ and $H_{j}$ for convenience.
In consequence, the objective \eqref{eq:obj} and gain functions \eqref{eq:gain} have to be adapted to consider gradient and hessian values from all classes.
A common approach in multi-variate regression and multi-target classification is to compute the average loss of the model over all targets \citep{waegeman19multitarget}.
Adapted to our XGBoost trees, this corresponds to the sum of $\frac{G_j^2}{H_j+\epsilon}$ over all labels (cf. Table~\ref{tab:splitmethods}). We refer to it as the \textbf{\ap{sumGain}} split method.
We use cross entropy as our loss, as it has demonstrated to be appropriate practically and also theoretically for binary and especially  multi-label classification tasks \citep{nam14revisiting,dembczynski12PCCdependence}. \comds{warum CE nicht für Multilabel loss angegeben?}
Hence, the loss is computed as (shown here only for a single label)
\begin{align}\label{eq:crossentropy}
    l_{ce}(y,\hat{y})=-y\log(\hat{y})+(1-y)\log(1-\hat{y}).
\end{align}
In order to get $\hat{y}$ as a probability between zero and one, a sigmoid transformations has to be applied to the summed up raw leaf predictions $\tilde{y}=\sum_{t=1}^{T} f_t(\vx)$, returned from all boosting trees, where $\hat{y} = \sigmoid(\tilde{y}) = \frac{1}{1+e^{-\tilde{y}}}$. This is also beneficial for calculating $g$ and $h$, since the gradients of the loss function simply become
\begin{align}\label{eq:gh}
    g = g_{ce} = \nabla_{\hat  y} l_{ce}(y, \hat y) = \hat{y} - y \hspace{0.5cm}\text{and}\hspace{0.5cm} h = h_{ce} = \nabla^2_{\hat  y} l_{ce}(y, \hat y) = \hat{y} \cdot (1-\hat{y}).
\end{align}
One might not expect a very different prediction from the combined formulation than from minimizing the loss for each label separately by separate models (as by BR). 
However, as \citet{waegeman19multitarget} note, fitting one model to optimize the average label loss has a regularization effect that stabilizes the predictions, especially for infrequent labels.
In addition, only one model has to be inferred in comparison to $N$, which has a major implication on the computational costs. 
This is especially an advantage in the case of a large number of labels and our proposed dynamic approach can directly benefit from it.\com{gefaellt mir nicht ganz letzter satz}

There are only few special adaptations of the gradient boosting approach to MLC in the literature and they mainly deal with computational costs.
Both \citet{pmlr-v70-si17a} and \citet{zhang2019GBDT-MO} propose
to exploit the sparse label structure which they try to transfer to the gradient and Hessian matrix by using $L$0 regularization.
These approaches are limited to decomposable evaluation measures (such as \eqref{eq:crossentropy}), which roughly speaking means that, opposed to the classifier chains approaches, they are tailored towards predicting the labels separately rather than jointly.
Moreover, different technical improvements regarding parallelization and approximate split finding are proposed which could also be applied to the proposed technique in the following.
Recently, \citet{rapp20boomer} proposed to use gradient boosting in order to induce classification rules. Instead of predicting the labels in sequence, the rules predict all labels at once, which allows for minimizing also non-decomposable losses. On the other hand, previous predictions can only be exploited indirectly.

\subsection{Gradient Boosted Dynamic Classifier Chains}
\label{sec:XDCC}
After introducing the ML-XGBoost models, which can deal with multiple labels, the next step is to modify the tree construction to align it with our goal of predicting in each round a single label per instance. 
Depending on the strategy of ordering the labels, different ways of constructing the tree might be necessary.
In our case, we adapt the tree construction process to the label ordering strategy by modifying the splitting criterion at the inner nodes. 

\begin{table}[tb]
\caption{Proposed split gain \comds{gibt auch noch die Bezeichung split function und split-method -> vereinheitlichen} calculations with a simplified example calculation for the predicted scores $\hat{\vy}=(0.8,0.2,0.9,0.1)$ of the previous trees  and given true labels $\vy=(1,1,0,0)$. For convenience, we  assume $H_j+\epsilon=1$.}
\resizebox{\textwidth}{!}{
\begin{tabular*}{\textwidth}{c @{\extracolsep{\fill}} c @{\extracolsep{\fill}} c @{\extracolsep{\fill}} c @{\extracolsep{\fill}} c @{\extracolsep{\fill}} c}
\hline
Gain & Formula & Example & Gain & Formula & Ex. \\
\hline
\ap{sumGain} & $\displaystyle\sum_{j=1}^N \left(\frac{G_j^2}{H_j+\epsilon}\right)$
& $0.2^2+0.8^2+0.9^2+0.1^2$ &
\ap{maxGain} & $\displaystyle\max_{1 \leqslant j \leqslant N}\left(\frac{G_j^2}{H_j+\epsilon}\right)$
& $0.9^2$ \\
\ap{sumSigned} & $\displaystyle\sum_{j=1}^N \left(\frac{-G_j}{H_j+\epsilon}\right)$
& $0.2+0.8-0.9-0.1$ &
\ap{maxSigned} & $\displaystyle\max_{1 \leqslant j \leqslant N}\left(\frac{-G_j}{H_j+\epsilon}\right)$
& $0.8$ \\
\ap{sumAbsG} & $\displaystyle\sum_{j=1}^N \left(\left|\frac{-G_j}{H_j+\epsilon}\right|\right)$
& $0.2+0.8+0.9+0.1$ &
\ap{maxAbsG} & $\displaystyle\max_{1 \leqslant j \leqslant N}\left(\left|\frac{-G_j}{H_j+\epsilon}\right|\right)$
& $0.9$ \\
\end{tabular*}
}
\label{tab:splitmethods}
\end{table}

Table~\ref{tab:splitmethods}
shows the proposed split functions and an example for each one to demonstrate the  calculations.
They replace the formulation of ${obj}^*$  in \eqref{eq:gain}.
In the example in the table, we assume to have a single instance with four different target labels $\vy \in [0,1]^4$ and their corresponding predictions $\hat{\vy}$. $g$ and $h$ are calculated according to \eqref{eq:gh} and we get $G=(-0.2,-0.8,0.9,0.1)$.
\raus{We have focused on different characteristics for each function. \com{Brauchen wir den Teil von hier bis ...}The \textit{max} versions focus on optimizing a tree for predicting only a single label, whereas \textit{sum} functions aim for finding a harmonic split that generates predictions with high probabilities over all labels. The \textit{weight} variants focus on directly optimizing the tree outputs and hence prefer positive labels, while the \textit{gain} splits stay close to XGBoosts original gain calculation and try to optimize positive and negative labels to the same extend. \com{... hier? oder auch das danach?}\res{der Teil zwischen den Kommentaren kann eigentlich raus. Unten in den drei Absätzen steht das ja eigentlich nochmal detaillierter}}
Hereinafter we give a more detailed description and motivation for each gain function:
\begin{description}
\item[\textbf{Maximum default gain over all labels.}]
XDCC predicts labels one by one. It hence does not need to find a split which increases the expected loss over all labels (such as \ap{sumGain}), but only one.
Hence, \textbf{\ap{maxGain}} is tailored to find the label with maximal gain, which corresponds to the label for which the previous trees produced the largest error.
In the example in Table~\ref{tab:splitmethods}, this corresponds to $\lambda_3$ for which a change of $0.9^2$ w.r.t. cross entropy was computed if the prediction is changed to the correct one.
\item[\textbf{Sum and maximum gradients over all labels.}]
~~ In contrast to \ap{maxGain},\linebreak  \textbf{\ap{\mbox{sumSigned}}} 
aims at good predictions for positive labels only and hence corresponds to the idea of predicting the positive labels first.
Positive labels obtain positive scores, whereas negative labels obtain negative scores. The variant \textbf{\ap{maxSigned}} chooses the  positive label for which the greatest improvement is possible and only goes for the best performing negative label if there are no true positive labels in the instance set.
In the example, $\lambda_2$ is chosen since the improvement is greater than for $\lambda_1$, and definitely greater as for the negative labels.
\item[\textbf{Sum and maximum absolute gradients over all labels.}] 
\qquad Different to  \mbox{\ap{sumGain}} and \ap{maxGain}, the measures \ap{sumSigned} and \ap{maxSigned} not only favour positive labels but also take the gradients linearly instead of quadratically into account.
This might, for instance,  reduce the sensitivity to outliers.
Hence, we also include two variants \textbf{\ap{sumAbsG}} and \textbf{\ap{maxAbsG}} which encourage to predict the labels where the model would improve the most, regardless whether it is positive and negative, but which similarly to \ap{sumSigned} and \ap{maxSigned} use a linear scale on the gradients.
\end{description}
Even though DCC's original design is to predict a single label per round,
good overall predictions might be required from the beginning for instance in the case of shorter chains.
Therefore, we use the split-method as an additional hyperparameter to choose it individually for different XDCC variants and datasets.\com{das am ende nochmal überprüfen, ob wir das brauchen}

\begin{figure}[tb]
    \centering
    \resizebox{\textwidth}{!}{\includegraphics{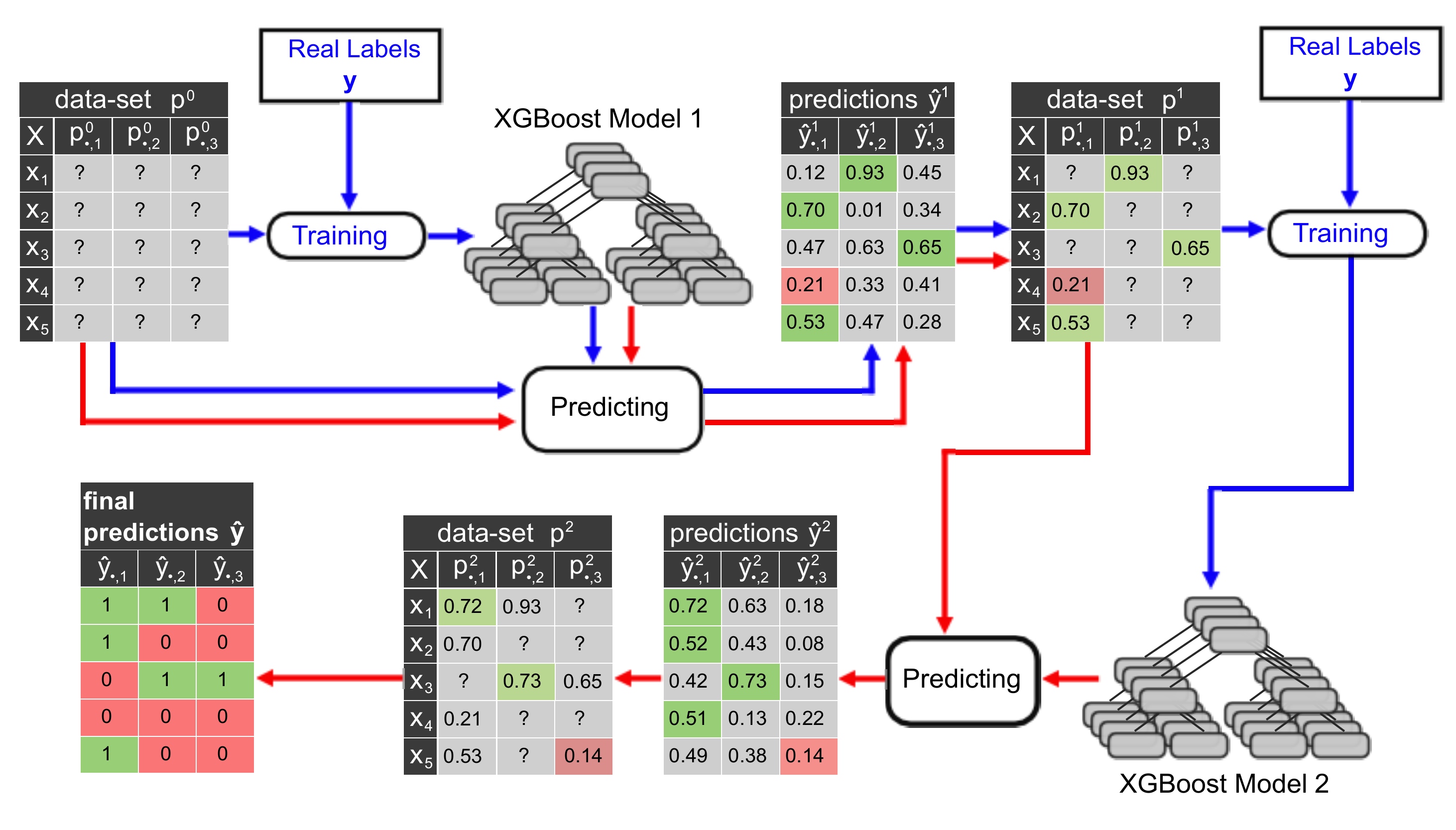}}
    \caption{Dynamic Chain: Example training pipeline (blue arrows) and prediction pipeline (red arrows) for a chain with length two that can predict up to two positive labels per instance.}
    \label{fig:chain-train}
\end{figure}

\subsubsection{Training Process}
As for classical classifier chains, and differently from RDT, XGB trains a separate classifier for each label prediction round.
The main difference to CC is of course again, that the next label to be predicted can be different for each instance, as it is chosen based on a dynamic prediction strategy. \com{maybe here refer to strategy section}
\raus{The process of constructing the sequence of label predictions is basically the same during training and testing, with the only difference that the classifier needs to be first build before being applied during the training process.
In addition, and similarly to \emph{nested} classifier chains \citep{senge13rectifyingCC}, we use the predicted instead of the true values for the labels in subsequent training rounds\com{hm, but why do we so}.
}%
A schematic view for training the dynamic chain with a length of two is shown in Figure~\ref{fig:chain-train} following the blue lines.

Similarly to RDT dynamic chain method in Section~\ref{sec:collapsedCC}, we initialize the augmented label features
$\vp$
of each train instance $\vx$ with ``?".
While proceeding through the chain, these ``?" values are     replaced with predicted label probabilities
out of prediction vector $\hat{\vy}^k  \in (0,1)^N$.
	As soon as these feature columns begin to be filled with values, following classifiers may detect dependencies and base their predictions on them.
	In each round $k$, for $1\le k \le N$, starts with training a new ML-XGBoost $f^{1,k}\ldots f^{T,k}$ model of $T$ trees by passing the train set combined with the additional label-features $\vp^{k-1}$ and the target label matrix $\vy$ to it.
	Afterwards, the model is used to generate predictions
    $\hat{\vy}^k=\sigmoid\left(\sum_{t=1}^{T} f^{t,k}((\vx,\vp^{k-1}))\right)$
	on the same data used to train it, shown in the \textit{predictions} tables.
In the last step these predictions are then propagated to the next chain classifier by replacing the corresponding label features $\pi_k$ chosen by the chaining strategy with the corresponding predicted probabilities,
i.e., $p^k_{\pi_k}=\hat{y}^k_{\pi_k}$.
\com{hier fehlt, daß wir predicted probabilities, wie in nested classifier chains \citep{senge13rectifyingCC}, und nicht die echten binären targets verwenden wie in CC. ist vermutlich nicht der wichtigste punkt}

\subsubsection{Dynamic Chain Ordering}
In order to be able to shorten the training and prediction process
we follow  \eqref{eq:DCCpipositives} and propagate the label with the highest estimated probability first.
However, we combine this ordering on the positive labels with RDT's strategy 
of selecting labels by their certainty \eqref{eq:DCCpicertainty} on the negative labels.
More specifically, we start to select the label with the lowest probability next as soon as no further label with probability higher than 0.5 is found.
The objective of this strategy is two-fold:
in case the prediction process stops before round $N$, it is desirable 1) to have returned as many positive labels as possible and 2) that the predictions so far are as accurate as possible.

The strategy can be formalized as follows where 	$J=\{1 \ldots N\}\backslash \{\pi_1 \ldots \pi_{k-1}\}$ denotes the labels which were not propagated previously:
    \begin{align}
	  \pi_r =
   \begin{cases}
     \argmax\limits_{j\in J} \hat{y}_{j}^k,    & \text{if }  \max\limits_{j\in J} \hat{y}_{j}^k \geq 0.5\\
     \argmin\limits_{j\in J} \hat{y}_{j}^k,    & \text{if }  \max\limits_{j\in J} \hat{y}_{j}^k < 0.5
   \end{cases}
	\end{align}
Note that in practice it can still happen that a positive labels is found  after a negative one, for instance because the negative predictions added evidence for a certain label to be relevant.
For the same reason certainties for already set labels in $\pi$ might also increase in subsequent rounds.
In order to benefit from these increased certainties, we allow to update the scores in $\vp$ in these cases.\com{elm: so habe ich das nun formuliert. es ist nicht ganz vollständig, aber die formel für $\pi$ stimmt, wenn man bei solchen probability update runden $r$ nicht hochzählt (so daß man am Ende weniger als $N$ elemente in $\pi$ hat)}
However, we do not allow that later classifiers revoke previous decisions by changing labels from positive to negative or the other way around.


\raus{
	Three different cases can occur during this process: 
	\begin{itemize}
	    \item At least one label, that was not propagated previously, has a probability $\ge 0.5$: The label with the highest probability is propagated.
	    \item All labels, that were not propagated previously, have probabilities $< 0.5$: The label with the lowest probability is propagated.
	    \item Otherwise, no additional label is propagated.
	\end{itemize}
They can be formalized where $p_{i,j}^k$ denotes the added label feature and $\hat{y}_{i,j}^k$ the corresponding predictions for label $\lambda_j$ of an instance $\vx_i$ in training round $r$.
    \begin{align}
	  p_{j}^k =
   \begin{cases}
     \hat{y}_{j}^k    & \text{if } p_{j}^{r-1} = ? \text{ and }  \max_m \hat{y}_{m}^k \geq 0.5 \text{ and }   \hat{y}_{j}^k = \max_m \hat{y}_{m}^k \\
     \hat{y}_{j}^k    & \text{if } p_{j}^{r-1} = ? \text{ and }   \max_m \hat{y}_{m}^k < 0.5 \text{ and } \hat{y}_{j}^k = \min_m \hat{y}_{m}^k \\
     p_{j}^{r-1}      & \text{otherwise }
   \end{cases}
	\end{align}
    \begin{align}
	  p_{i,j}^k =
   \begin{cases}
     \hat{y}_{i,j}^k    & \text{if } p_{i,j}^{r-1} = ? \text{ and }  \max_m \hat{y}_{i,m}^k \geq 0.5 \text{ and }   \hat{y}_{i,j}^k = \max_m \hat{y}_{i,m}^k \\
     \hat{y}_{i,j}^k    & \text{if } p_{i,j}^{r-1} = ? \text{ and }   \max_m \hat{y}_{i,m}^k < 0.5 \text{ and } \hat{y}_{i,j}^k = \min_m \hat{y}_{i,m}^k \\
     p_{i,j}^{r-1}      & \text{otherwise }
   \end{cases}
	\end{align}
    In all cases where labels are propagated, later classifiers are not allowed to change these labels from positive to negative or the other way around, based on the assumption that later classifiers tend to have a higher error rate, since their decisions are based on previous predictions \citep{senge2014problem}.
}

	\subsubsection{Prediction Process}
	\label{sec:predictionprocess}
	The prediction process is similar to the training process. Instead of training a model in each step, we reuse the models from the training phase to generate predictions on the test set. After all predictions are propagated, the propagated labels are mapped to label predictions, where probabilities
$p_{j} < 0.5$ or equal to ``?" are interpreted as negative labels and probabilities $p_{j} \ge 0.5$ as positive labels. The process is depicted in Figure~\ref{fig:chain-train} following the red lines.

\skp{	
	\subsection{Refinements to the chain}\tod{wenn noch zeit ist integrieren (seco wird 5.3.3 etc)}
	In this section we shortly describe two problems we faced during development of the DCC approach and propose two crucial methods to tackle them.
}
\subsubsection{Separate and Conquer}
We faced the following problem during the adaptation  of the DCC approach to XGBoost.
	Consecutive models in the chain tend to select the same splits and therefore predict the same labels, especially ones which are easy to learn, e.g. if they clone existing features. We solve this problem by introducing an approach similar to separate-and-conquer from rule learning \citep{jf:AI-Review} that is applied after each feature column update\rev{, i.e., after learning $f^{1,k}\ldots f^{T,k}$ and as preparation for learning $f^{1,k+1}\ldots f^{T,k+1}$}. The \textit{separating} step turns all gradient and hessian values of previously predicted labels for an instance to zero. Thereby, they are no longer considered during split score calculation in the \textit{conquering} step and other splits become more likely since scores for already used splits are lower.
\rev{The computation of the prediction scores $w_v$ at the leafs is not affected by this measure
so that labels still get the chance to be selected as next label for instances for which they were not yet chosen.
This aspect is also relevant for the following measure.
}

\subsubsection{Cumulated Predictions}
A second observation during development was that final predictions, after traversing the chain, contain too little positive labels. Analyzing the chain models showed that especially early models predict multiple positive labels, but are only allowed to propagate the one with the highest probability. Therefore we introduce \textit{cumulated predictions} to preserve these otherwise forgotten positive predictions.
	The idea is to save all predictions of each chain classifier and merge them afterwards with the chain predictions of the unmodified DCC using the following heuristic.
	The final cumulated prediction $c_{j}$ for label $\lambda_j$ and an instance $\vx$ is computed as
    \begin{align}
	  c_{j} =
   \begin{cases}
    p_{j}^N                                   &\text{if $p_{j}^N \ne$ ?} \\
    \max(\hat{y}_{j}^1, ..., \hat{y}_{j}^N)      &\text{otherwise}
   \end{cases}
	\end{align}
\rev{
These final predictions $c_{j}$ are used as in Section~\ref{sec:predictionprocess} in order to determine whether the label is set or not.
For example, the standard version predicts $\lambda_1$ as negative for test instance $\vx_4$ in Figure~\ref{fig:chain-train} since the label was chosen as next label in the first round and set to negative due to its probability of $\hat{y}^1_1=0.21$.
In contrast, the cumulative approach would set $\lambda_1$ as relevant since the second chain model predicted a probability of $\hat{y}^2_1=0.51$.
}

\section{Experiments \com{2-3p}}
\label{sec:XGBevaluation}
The purpose of the experimental evaluation is two-fold.
Firstly, we want to directly compare the proposed dynamic extension of gradient boosting trees to the static classifier chain variant, both in terms of predictive performance and computational costs (Sections~\ref{sec:xgb_CCvsDCC}, \ref{sec:comparisonXGB}).
Secondly, we present a comparison to established decision tree learners, including the random decision trees proposed above, in order to assess
 the practical implications of the proposed extensions (Section~\ref{sec:comparisonbaselines}).
In particular, we evaluated the following algorithms:
\begin{itemize}
\item \textbf{J48}: WEKA's implementation of C4.5 represents in our comparison the family of classical single decision tree learners\raus{ which address a learning task by choosing splits at the inner nodes which optimize a certain pre-determined criterion such as the information gain}.
\item \textbf{BR}: Binary relevance learning, which learns on J48 decision tree for each label.
\item \textbf{CC}: Classifier Chains, which extend BR by including previously predicted labels as additional features for subsequent labels.
\item \textbf{LP}: The label powerset algorithm, which treats every label combination as a separate class value for J48.
\item \textbf{RF}: Random forest (and its variant predictive clustering trees) regularly achieve best positions in comparisons of state-of-the-art algorithms for multi-label classification \citep[no comparison to gradient boosted trees, though]{madjarov12MLCcomparison,bogatinovski21MLCcomparison}. 
We used the WEKA implementation of random forests as base learner for BR, CC and LP.
\item \textbf{XGB}: XGBoost  used as base learners for BR and CC.
\item \textbf{RDT-DCC}: Dynamic classifier chains using random decision trees (Section~\ref{sec:dynamicpredictions}).
\item \textbf{ML-XGB}: A single multi-label XGBoost model introduced in Section~\ref{sec:ML-XGB}. 
\item \textbf{XDCC\textsubscript{cum}}: DCC with ML-XGB models as base classifiers, as proposed in Section~\ref{sec:XDCC}, and cumulated predictions turned on.
\item \textbf{XDCC\textsubscript{std}}: The variant of XDCC  without cumulated predictions, included in order to show the effect of this modification.
\end{itemize}
Hyper-parameters, especially regarding the tree construction, were optimized for F1 on a randomly
selected 20\% subset of the training set which was fixed beforehand.%
\footnote{
The following parameters were tuned by grid-search:
Number of trees \{100, 300, 500\},
max. tree depth \{5, 10, 30, 50\}
for RF;
Number of trees \{100, 300, 500\},
max. leaf size \{3, 5, 9\}, 
max. tree depth \{5, 10, 20, 30, 50\},
percentage of label tests \{0.1, 0.2, 0.3\}
for RDT;
Max. tree depth \{5, 10, 20, 50, 100\},
number of boosting rounds \{10, 20, 50, 100\},
learning rate \{0.1, 0.2, 0.3\} for XDCC, ML-XGB, XGB-BR, XGB-CC;
Split methods in Table~\ref{tab:splitmethods} for XDCC, ML-XGB.
All remaining parameters were set to default values.
}
\com{include here the cross optimizations? For some comparisons between the XGBoost models, we the best parameters found for a different variant. / For shortening the process, we usually used the best parameters obtained by ML-XGB for configuring the XDCC variants. Unless otherwise stated, a a fixed random chain was used for CC.
}
This setting provided more stable results than using subset accuracy as the objective measure especially for the larger datasets.

\begin{figure}
    \centering
    \scalebox{.48}{\input{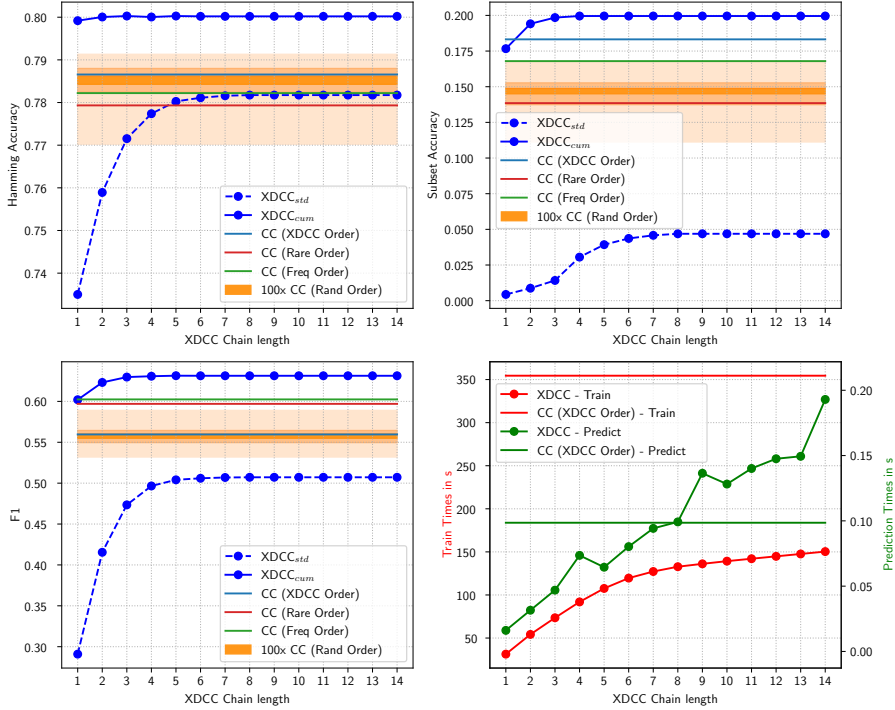}}%
    \caption{Comparison with respect to length of the chain on \ds{yeast}.
    XDCC-Order is overlayed by Rand-Order for F1. \rev{The shaded area for CC depicts the 5-quantiles (highest value, 20th, 40th, 60th, 80th value, lowest value).} }
    \label{fig:yeast}
\end{figure}

\raus{
\begin{figure}
    \centering
    \scalebox{.48}{\input{figures/tmc2007.pgf}}%
    \caption{Comparison with respect to length of the chain on \ds{tmc2007}. 
    }
    \label{fig:tmc}
\end{figure}
}

\subsection{Comparison of Split Functions}

\begin{table}%
\centering
\caption{Average ranks over the 11 datasets (and ranks over these in brackets) of the split function methods in combination with the cumulated and standard method.}
\label{tab:gaincomparison}
\rev{
\begin{tabular}{llccc}
\toprule
Variant & Gain	&	\hspace{2em}HA\hspace{2em}		&	SA		&	F1		\\
\midrule
XDCC\textsubscript{cum}
&	\ap{sumGain}	&	5.00	(2)	&	4.50	(1.5)	&	4.46	(2)	\\
&	\ap{sumSigned}	&	9.59	(11)	&	10.00	(11)	&	9.59	(11)	\\
&	\ap{sumAbsG}	&	7.09	(10)	&	6.23	(6)	&	5.18	(4)	\\
&	\ap{maxGain}	&	4.27	(1)	&	4.50	(1.5)	&	4.36	(1)	\\
&	\ap{maxSigned}	&	5.46	(5)	&	5.41	(4)	&	5.55	(5)	\\
&	\ap{maxAbsG}	&	5.91	(7)	&	5.55	(5)	&	4.64	(3)	\\
\midrule
XDCC\textsubscript{std}
&	\ap{sumGain}	&	6.59	(8)	&	6.32	(8)	&	7.18	(10)	\\
&	\ap{sumSigned}	&	10.55	(12)	&	10.64	(12)	&	11.41	(12)	\\
&	\ap{sumAbsG}	&	7.05	(9)	&	6.96	(10)	&	6.64	(8)	\\
&	\ap{maxGain}	&	5.41	(4)	&	5.32	(3)	&	6.46	(7)	\\
&	\ap{maxSigned}	&	5.23	(3)	&	6.27	(7)	&	5.55	(5)	\\
&	\ap{maxAbsG}	&	5.86	(6)	&	6.32	(8)	&	7.00	(9)	\\
\bottomrule
\end{tabular}
}
\end{table}

\rev{
In a first step, we compared the proposed split gain functions.
The average ranks in each column in Table~\ref{tab:gaincomparison} were obtained by optimizing hyper-parameters for the respective loss.\footnote{The Friedman test passed at $\alpha=0.05$ and the Nemenyi critical distance is 5.18.}\tod{einfach so lassen oder noch eine erklärung, wieso wir das sonst nicht so gemacht haben (zu lange computation für CC?).}

Our expectation was that the \ap{max} heuristics, which finds splits leading to confident predictions for only one single label, should work particularly well with the proposed dynamic approach, since it fits to the idea of predicting labels one by one.
In fact, the results show that the \ap{maxGain}, \ap{maxSigned} and \ap{maxAbsG} methods generally beat their \ap{sum} counter-parts in direct comparison.
In particular \ap{maxGain}, which is based on the idea of predicting the most confident labels first, is the best method (though the better methods are generally close together) also when considering the standard and cumulative version of XDCC separately.
The \ap{maxSigned} criterion, which tries to fulfill the requirement of predicting positive labels first, performs worse.
However, the direct comparison between both methods w.r.t. F1 and XDCC\textsubscript{cum}, where \ap{maxGain} won on 7 and \ap{maxSigned} on 4 datasets, suggests that the best criterion depends to a great extent on the task at hand.
Therefore, we included the selection of the appropriate splitting criterion as an additional hyper-parameter to be optimized in the following experiments.

The \ap{maxAbsG} variants, which were meant to be less sensitive to outliers in the gradient computation, performed quite comparable to the base \ap{maxSigned} variant.
The bottom ranks of the \ap{sumSigned} heuristic, especially in comparison to the much better \ap{sumAbsG} variant, suggest that outliers are less problematic if the split is targeted to separate one single label where a confident prediction is possible.

}	

\subsection{Static vs. Dynamic Label Orderings}
\label{sec:xgb_CCvsDCC}

Similarly to Section~\ref{sec:RDTevaluation}, we performed experiments which allow to see the advance of the DCC approach obtained by subsequently refining its predictions.
Note, however, that contrary to the evaluation w.r.t. RDTs we cannot fully isolate the comparison between static and dynamic predictions from other effects
than the order of the chain.
Hence, we included an instantiation of CC's static chain which  was generated by combining the
dynamic chains  found by XDCC for the instances in the validation set (XDCC Order).
This ensures a certain proximity between the static and dynamic models and hence further cancels out, to a certain degree, effects caused by the random selection of the static orderings.
Moreover, we included \rev{static but randomly ordered chains} (Rand Order), and the static orderings from  rare to frequent labels (Rare Order) and vice-versa (Freq Oder) as proposed by \citet{jn:NIPS-17-MLC-RNN}.
Each model used its own best hyper-parameters for the comparison of the predictive performance. However, XDCC used the XGBoost parameters found for CC for the comparison of the computational costs in order to obtain tree models of similar size.

As described in Section~\ref{sec:XDCC}, XDCC 
can provide useful predictions after each round.
This allows to terminate the prediction process early,
which can be a major advantage over CC in terms of computational costs.\com{elm: gilt das auch für die std-version? sb: Im Prinzip schon. Es ist nur so, dass die std Version in Runde m höchstens auch m Label vorhersagen kann. elm: aber wie wird es hier gemacht? also wenn label i nach runde r noch nicht vorhergesagt wurde, was wird dann für i vorhergesagt wenn nach r abgebrochen wird?}
Moreover, by subsequently refining its predictions based on previous predictions,
we did not only expect to advance regarding F1, for which we optimized the XGBoost parameters, but especially in terms of SA.\tod{diesen satz ist natürlich für tmc gefährlich, aber vlt. kann man mit der niedrigen cardinality argumentieren}
Figure~\ref{fig:yeast} shows measures HA, SA and the time for training for different lengths of the chain in full detail exemplarily on \ds{yeast}.
The previous experiments on RDT (Section~\ref{sec:RDTevaluation}) showed that this dataset is appropriate to investigate the effect of static  vs. dynamic label orderings, as it has a relatively high cardinality and appears to possess a label dependence structure which can be exploited by chaining techniques.
\rev{
The relatively small size of the dataset also allowed us to repeat CC 100 times in order to show the range of results if the permutations are chosen purely randomly.
}
The trade-off between predictive performance and computational costs for the remaining datasets is analyzed further below.
Note that length one of XDCC$_{cum}$ corresponds to ML-XGB when the same parameter were used.\tod{Welche Parameter verwendet XDCC-cum und welche XDCC-sum? Offenbar nicht die gleichen, weil sonst müssten sie vom selben Punkt aus starten? Was wäre die Performance von ML-XGB?} \res{Doch, es sind die gleichen Parameter. Der unterschied ist, dass ML-XGB im Prinzip die Cum Variante mit Länge 1 ist, d.h. nachdem von Model 1 pro Instanz ein Label vorhergesagt wurde, werden  die Predictions aller Modelle (was in dem Fall nur das eine ist) mit reingemerget. Bei STD findet das merging nicht statt und es gibt bei Kette Länge 1 höchstens auch 1 positives Label pro instanz}

\begin{figure}[t]
    \centering
    \resizebox{0.9\textwidth}{!}{\includegraphics[trim={0cm 0.0cm 0cm 0.0cm},clip]{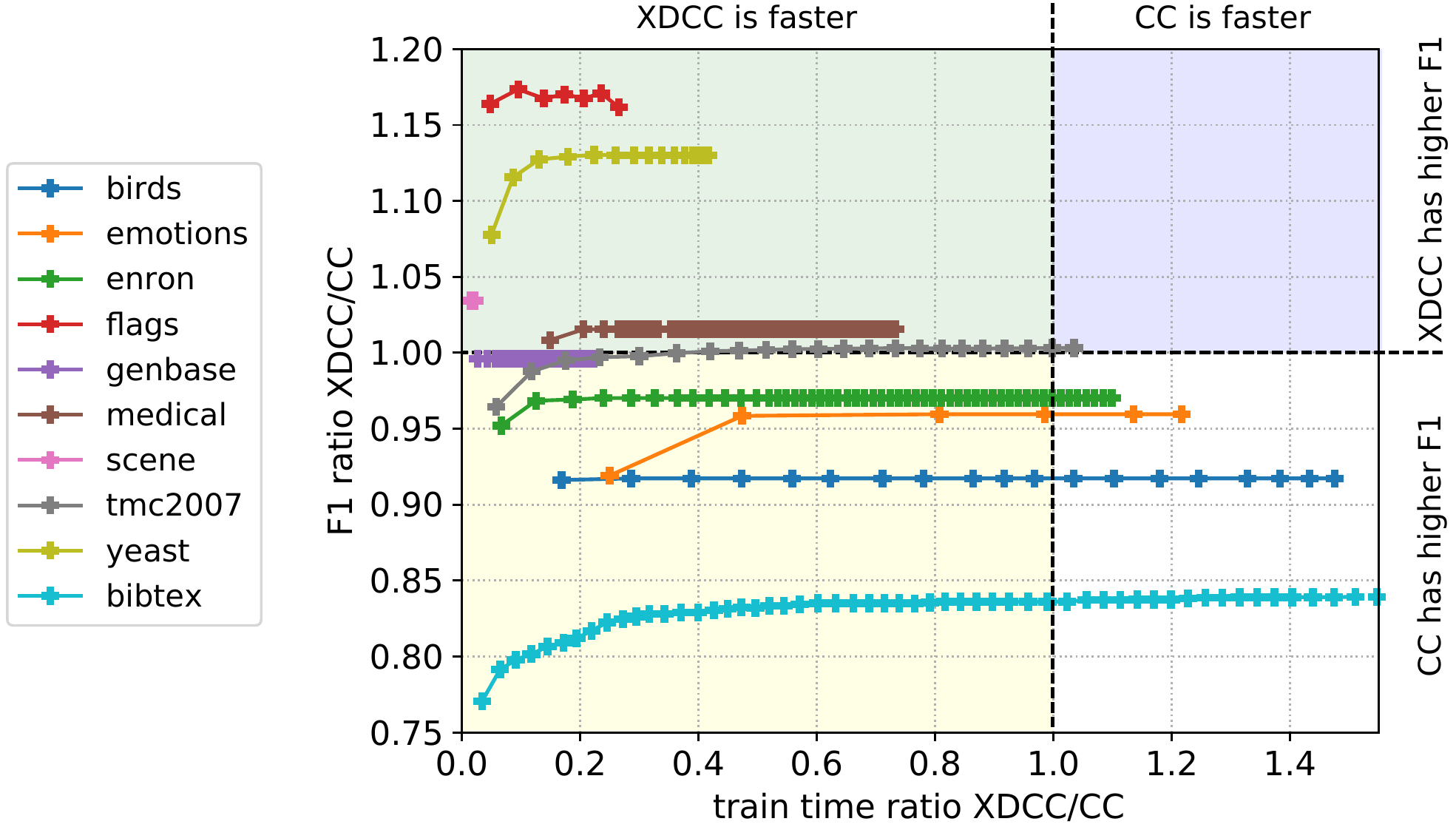}}
  \caption{Train time ratios between XDCC$_{cum}$ and CC in relation to their ratio with respect to F1 for nine datasets.
  Both models use individual XGB parameters where they performed best.
  \ds{CAL500} around (0.1,1.4) not shown for convenience, \ds{bibtex} continues to (5.3,0.84).
  \com{zur erinnerung: XDCC mit parameters optimiert für ML-XGB, CC mit random order mit parametern optimiert für CC}}
  \label{fig:ratiosSelfTuned}
\end{figure}

The first observation in the two graphs in the top row is that, as expected, in terms of HA and SA, the performance of standard XDCC, which makes one prediction per label, improves with increasing length until a little bit further than the average cardinality of four of the dataset.
If we add the cumulated predictions, the performances converge much faster because we can make more predictions in each iteration. Yet, there is a clear improvement visible for SA, which indicates that XDCC$_{cum}$ is able to directly benefit from the previous predictions in order to match the correct label combinations.
The cumulated predictions are also decisive for surpassing the static CC chains. 
\rev{This includes the 100 random ordering as well as the orderings according to the label frequencies and XDCC's ordering.}
Interestingly, as can be seen from the red curves in the lower right graph, the training costs of CC 
are never reached although the same XGBoost parameters were used. In terms of prediction costs, XDCC becomes more expensive after eight rounds (green curve), which is long after it has reached the average cardinality and XDCC's optimal predictive performance.

\raus{
The results for \ds{tmc2007} depicted in Figure~\ref{fig:tmc} are more diverse and show that
the performance also depends on the dataset and objective measure chosen.
In terms of \emph{F1} (lower left graph), the performance development of XDCC is quite similar as in Figure~\ref{fig:yeast}, as expected by the target of the hyper-parameter optimization, with the difference that in this case, XDCC$_{cum}$ does not surpass CC. 
More surprising is that, in the case of HA (upper left), the optimal performance is reached after a single iteration of XDCC$_{cum}$, and that additional iterations even harm its predictive performance.
\tod{haben wir eine erklärung hierfür? @simon, kannst du was zum verlauf von precision und recall sagen?}
In terms of efficiency (lower right graph), we can again see that XDCC can be more efficiently trained, as it can be stopped after a few iterations (even after one or two iterations in the case of HA). Also note that on this dataset, the randomly chosen order for CC outperformed all other fixed orders, which may indicate a particularly lucky choice here.
Optimizing a fixed chain, especially by trying out several different orderings, would multiply the training time by the number of different orders tried and thus  be even more costly (cf.\ Section~\ref{sec:CC}).
}
\raus{
This is different on \ds{tmc2007}, as depicted in Figure~\ref{fig:performance-tmc}.
On this dataset, predicting 3--4 rounds of XDCC takes as long as predicting all the labels in CC.
However, note that each instance is associated to around two positive labels on this dataset.
We can also observe that although the improvement with respect to subset accuracy along the chain is not sufficient to reach the performance of CC in this case, adding rounds leads to an improvement in terms of F1, which can be considered as trade-off between the two extreme measures SA and HA.
}

\raus{
\begin{figure}[b!]
    \centering
    \resizebox{0.9\textwidth}{!}{\includegraphics[trim={0cm 0.1cm 0cm 0.25cm},clip]{figures/F1_trainTime_ratios_XDCC-CC_same_parameters_v2.pdf}}
  \caption{\raus{Train time ratios between XDCC$_{cum}$ and CC in relation to their ratio with respect to F1 for nine datasets.
  Both models use the same XGB parameter set which was optimized for XDCC and F1.
  CAL500 starting at (0.03, 5.87) and ending at (0.50, 6.09) not shown for convenience, no results for \ds{bibtex}.
  \com{zur erinnerung: XDCC mit parametern optimiert für XDCC, CC  mit random order mit parametern optimiert für XDCC}}}
  \label{fig:ratiosXDCCtuned}
\end{figure}
}

\begin{figure}[t!]
    \centering
    \resizebox{0.9\textwidth}{!}{\includegraphics[trim={0cm 0cm 0cm 0cm},clip]{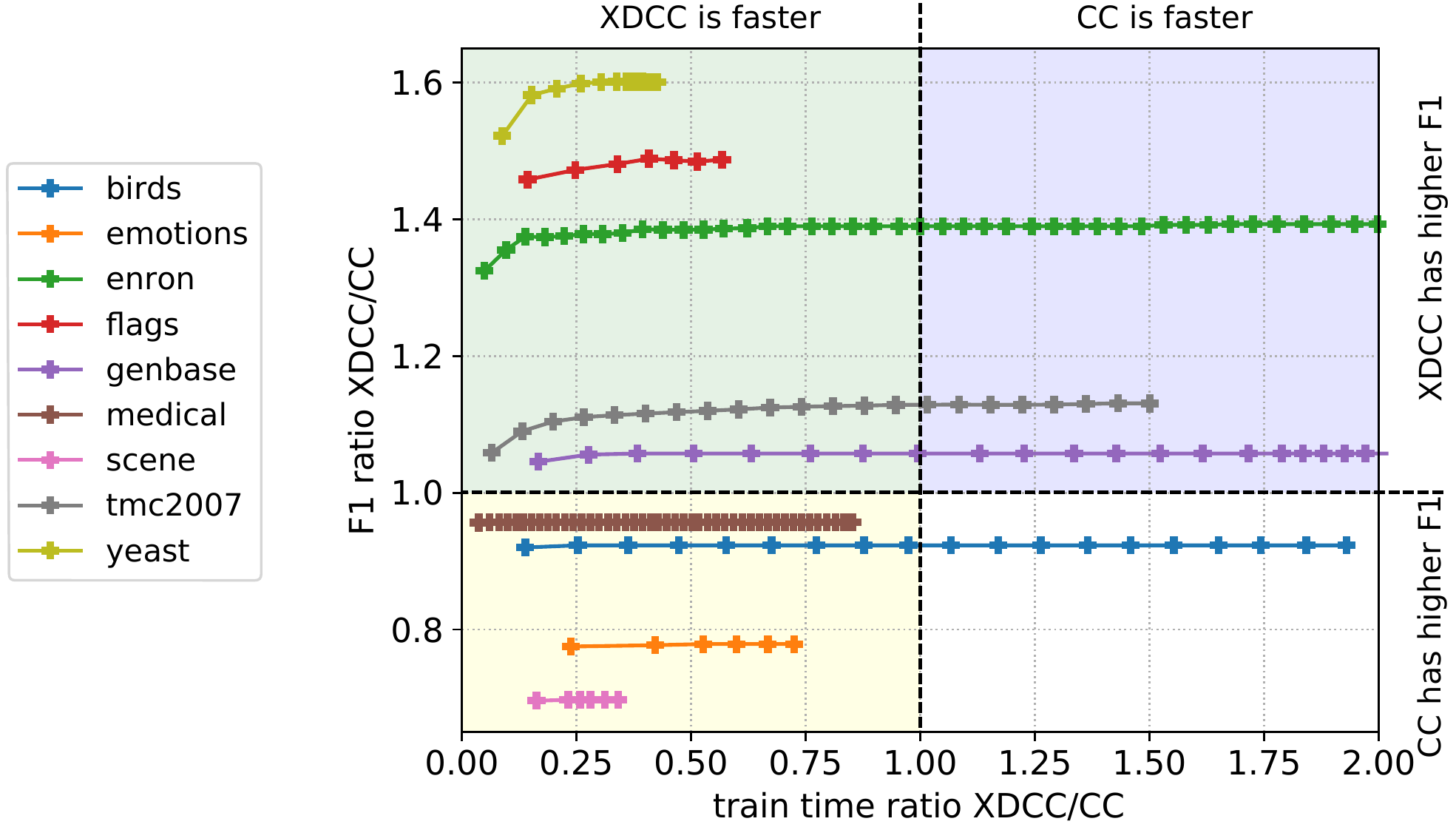}}
  \caption{Train time ratios between XDCC$_{cum}$ and CC in relation to their ratio with respect to F1 for nine datasets.
  Both models use the same XGB parameter set which was optimized for CC and F1.
  CAL500 starting at (0.008,  4.683) and ending (0.854, 4.690) not shown for convenience, no results for \ds{bibtex}.
  \com{zur erinnerung: XDCC mit parametern optimiert für CC, CC mit random order mit parametern optimiert für CC}
  }
  \label{fig:ratiosCCtuned}
\end{figure}

The point where train times of CC are reached by XDCC are further investigated in Figures~\ref{fig:ratiosSelfTuned} and \ref{fig:ratiosCCtuned}.
It shows the ratio of XDCC$_{cum}$  to CC  for the different datasets (connected lines) and chain lengths.
The difference between the two diagrams lies in the parameters used:
XDCC was trained with the best parameters found for CC in Figure~\ref{fig:ratiosCCtuned}.
In contrast, CC and XDCC had separate hyper-parameter optimizations in Figure~\ref{fig:ratiosSelfTuned}.
Hence, Figure~\ref{fig:ratiosCCtuned} is better suited for comparing the training times, whereas Figure~\ref{fig:ratiosSelfTuned} allows for a better comparison of the reachable predictive performance.
No clear general advantage was observed for neither of the frequency based label order heuristics in the previous analysis, especially not in comparison with a random order. Hence, we adopted static random chain orders in the following experiments.
Note that optimizing a fixed chain, especially by trying out several different orderings, would multiply the training time by the number of different orders tried and thus  be even more costly (cf.\ Section~\ref{sec:CC}).


We can observe in both diagrams that XDCC is always faster in the first rounds than CC.
Only for higher number of rounds the ratio is advantageous for CC, but at that point XDCC has always already converged w.r.t. predictive performance and additional rounds do not have a great impact.
As already seen in Figure~\ref{fig:yeast} \raus{and \ref{fig:tmc}} for \ds{yeast}\raus{ and \ds{tmc2007}},
we can also observe for the other datasets a steep increase in \emph{F1} in the first rounds which decelerates approximately when reaching the average number of positive labels per example.
Consequently, the graphs for \ds{scene}, \ds{birds}, \ds{genbase} and \ds{medical} with a cardinality around one are straight or only exhibit an increase in the very beginning.
Interestingly, the datasets for which XDCC has greater difficulties w.r.t. \emph{F1}  are also the ones where XDCC has to invest more time  than CC to learn the complete sequence.
Except for \ds{CAL500}, \ds{medical}, \ds{emotions}, the ordering of the endpoints
seems to correlate quite well with the density of the datasets, i.e., the cardinality divided by the total number of labels.
We leave further investigations of this relation for future work.

\raus{
XDCC starts with a relative short training time
Figure~\ref{fig:ratiosSelfTuned} shows that XDCC only takes longer than CC on four datasets and only for the last rounds.
For four of these datasets XDCC does not reach CC's F1.
As already shown in Figure~\ref{fig:performance-yeast}, XDCC$_{cum}$ only improves in the  first rounds, and sometimes there is even a tendency to decrease.
}

\begin{figure}[tb!]
    \centering
    \resizebox{\textwidth}{!}{
    \includegraphics{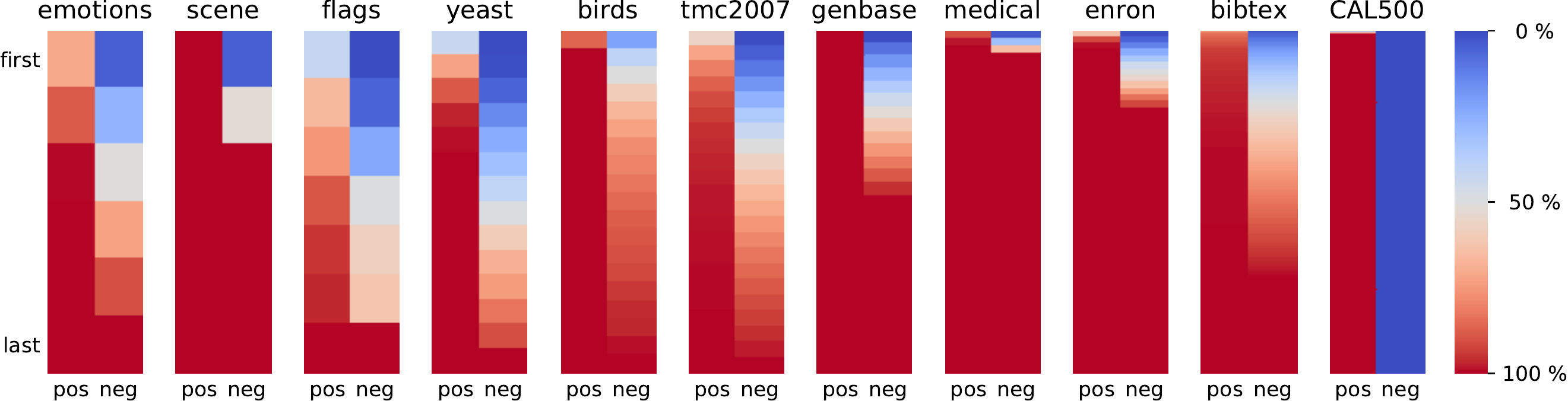}
    }
    \caption{Heat maps of the development of the predictions of positive  and negative labels (left and right side of the bar, respectively) from the first (top row) to last round (bottom row) given as fraction (color level) of the total number of positive and negative predictions on the respective dataset. 
    }
    \label{fig:prediction_development}
\end{figure}

The progress of predicting the labels is also depicted in Figure~\ref{fig:prediction_development}. 
The fast completion of the prediction of positive labels, as visualized by the achievement of 100\% on the respective left sides of the bars, indicates that  positive labels are generally predicted in earlier rounds, as expected from the design of the split functions.
As shown previously, this behaviour is decisive for the fast convergence and hence the possibility to end the training and prediction processes already in early rounds.
\tod{@simon: bei der vorhersage der negativen labels, also speziell wieso sie vorher aufhören können, kannst du vielleicht was sagen. so wie ich es verstanden habe, können mehrere labels in einer runde vorhergesagt werden. bedeuten hier die 100\% dass das passiert, und es kommen am ende keine weiteren negativen labels hinzu (algo wäre also eigentlich fertig weil alle N labels vorhergesagt).}
\tod{folgendes wieder reinnehmen, sobald wir zb eine zufriedenstelligere erklärung haben. das hauptproblem ist, dass wir den effekt nicht gut beschreiben können, da dafür die notwendigen algorithmischen details fehlen. also, die sind im paper nicht ausführlich beschrieben.}
\raus{
\ds{CAL500} seems to be a special case, since apparently the prediction process gets stuck after predicting only two positive labels. XDCC neither predicts positive nor negative labels after that point. \tod{@simon, könntest Du hier noch ein erklärungsversuch machen? vielleicht nehmen wir die ganze erklärung zu cal500 raus, aber dann haben wir wenigstens schon was in der hand falls rückfragen kommen}
}

In summary, on the analyzed datasets, XDCC's label ordering strategy allows the prediction process to terminate advance, resulting in a substantial speed-up in comparison to CC, without any major loss in predictive performance.

\subsection{Comparison to Decomposition Methods}
\label{sec:comparisonXGB}
Table~\ref{tab:XGBcomparison} summarizes the comparison between XDCC$_{cum}$ and CC (random chain order) but also includes a binary relevance model trained with XGBoost and the other XDCC variants.
Except for ML-XGB, which corresponds to stopping XDCC$_{cum}$ after the first round, the XDCC models processed the full chain for both training and prediction.\tod{@simon: fehlt hier noch was als hinweis zum setup? alle verfahren wurden bezüglich sich selbst optimiert, oder? bzw. wie war das mit ML-XGB? wurde XDCC nich bezüglich ML-XGB optimiert, oder andersrum (XDCC nach XDCC aber nach runde 1 abgebrochen)? ist XDCC-std einfach XDCC-cum (also auch so optimiert) aber mit unterschiedlichem vorhersageprozess?}
XDCC$_{std}$ is  included for
showing the effect of cumulative predictions.
\footnote{The Friedman test passed at $\alpha=0.05$ for all measures but the critical distance of 1.66 is only reached for some comparisons to XDCC$_{std}$.\com{vlt. fällt jemandem hier eine erklärung ein, warum das nicht so schlimm ist, z.b. ohnehin sehr ähnliche varianten und es geht nur um filigrane unterschiede.}}

The first observation is the strong baseline achieved by BR regarding HA, as partially expected from Section~\ref{sec:measures}.
In the same way, CC is best in terms of SA.
However, ML-XGB performs second regarding HA and XDCC$_{cum}$ is second regarding SA, which suggests that the proposed approach is able to trade-off between both extremes.
This is also confirmed by the best position in terms of F1.
The positive effect of the cumulative predictions is clearly visible by the direct comparison between both XDCC.
The comparison of the training and prediction times suggests that a similar advantage to BR is achievable as to CC when shortening the prediction process.

\begin{table}[tb]
  \caption{Predictive performance and times comparison of the XGBoost variants. Shown are the average ranks over the eleven datasets and the ranks over these in brackets.\tod{If time, make ranks of ranks smaller and put more space between columns.}}
  \label{tab:XGBcomparison}
\centering
    \begin{tabular}{lccccc}
\toprule
Method & \hspace{2em}HA\hspace{2em} & \hspace{2em}SA\hspace{2em} & \hspace{2em}F1\hspace{2em} & \hspace{0em}Train time\hspace{0em} & \hspace{0em}Test time\hspace{0em} \\
\midrule
XGB-BR	& 	2.10	(1)	& 	3.10	(4)	& 	2.60	(2)	& 	2.90	(2)	& 	2.46	(2)	\\
XGB-CC	& 	2.95	(3.5)	& 	2.25	(1)	& 	2.80	(3)	& 	3.50	(3)	& 	2.55	(3)	\\
\midrule
ML-XGB	& 	2.80	(2)	& 	2.85	(3)	& 	2.85	(4)	& 	1.20	(1)	& 	1.36	(1)	\\
XDCC$_{cum}$	& 	2.95	(3.5)	& 	2.60	(2)	& 	2.05	(1)	& 	3.70	(4.5)	& 	4.32	(4.5)	\\
XDCC$_{std}$	& 	4.20	(5)	& 	4.20	(5)	& 	4.70	(5)	& 	3.70	(4.5)	& 	4.32	(4.5)	\\
\bottomrule
    \end{tabular}
\end{table}

\subsection{Comparison to Baselines}
\label{sec:comparisonbaselines}

Table~\ref{tab:overallcomparison} presents a comparison of the proposed tree-based dynamic classifier chain approaches to baselines based on the J48 tree learner  and random forests.
As different technical infrastructure are employed, we do not include comparisons of the computational costs. \ds{tmc2007} did not complete for J48 and RF on time and was therefore excluded from the comparison. Classifier chains used random static orderings.\footnote{The Friedman test passed for HA and SA at $\alpha=0.05$, the critical distance is 3.05.}

Regarding Hamming accuracy, XDCC$_{cum}$ achieved the highest average rank even though it is not targeted at making correct individual label predictions, as the comparison to the BR decomposition using XGBoost demonstrated previously.
The more advanced techniques of gradient boosting seem to play out their advantage in this case.
XDCC$_{cum}$ also achieves good results on SA, though the label powerset method of RF clearly outperform the remaining algorithms on this measure.
\com{integrate?: Of particular interest is the comparison to the RDT-LP method. In terms of subset accuracy this method could outperform the dynamic chain methods on almost all datasets. Especially the results on the datasets \ds{Emotions}, \ds{Yeast} and \ds{Enron} are much better than the results of the other methods. However, a closer examination reveals that the results for the micro-averaged F1 measure of the RDT-LP method are not always that good. The label method could achieve a much higher score for micro-averaged F1 measure on the datasets \ds{Birds}, \ds{CAL500} and \ds{Enron}. \citet{senge2014problem} observed that LP can benefit from the restricted set of label combinations it can choose from, especially when the number of distinct combinations is relatively low, as it is the case for the used datasets. The other approaches, instead, have to make up valid combinations by concatenating single decisions. Whereas these single decisions might be better than for LP, as seen in terms of micro-averaged F1 measure, the complete combination might still be wrong especially if the cardinality is high.
}
\citet{senge2014problem} already showed that the LP method might be quite strong  when only a small fraction of the $2^N$ possible label combinations are observed in practice (or when the absolute number is generally low).
They argue that approaches like BR or CC have to make up valid combinations by concatenating single decisions whereas LP can stick to combinations for which there is certainly evidence.
Though these single decisions might be better than for LP, as seen in terms of Hamming accuracy, the
probability that the full combination is valid decreases exponentially with $N$.
Five of our dataset contain less than 100 distinct label combinations, and 7 less than 200, which might explain the good performance of LP.
\com{include ref to BOOMER?: For this reason, \citet{rapp20boomer} propose to map back to seen label combinations in order to allow for a fair comparison to LP.}
The advantage is taken over to F1, where RF-LP shares the first position  with our proposed XDCC$_{cum}$, but not to HA.
As aforementioned, the RDT has difficulties regarding sparse dataset, which is the reason for the low performance in direct comparison to the more broadly purposed baselines.
However, RDT-DCC surprisingly beats most of the baselines for F1.

More detailed results with raw performance scores for all the approaches and measures can be found in Tables~\ref{tab:ha_results}, \ref{tab:sa_results} and \ref{tab:f1_new} in the appendix.

\begin{table}[tb]
  \caption{Predictive performance comparison to the baselines. Shown are the average ranks over 10 datasets (all except \ds{tmc2007}) and the ranks over these in brackets. }
  \label{tab:overallcomparison}
\centering
    \begin{tabular}{lccc}
\toprule
Method & \hspace{3em}HA\hspace{3em} & \hspace{3em}SA\hspace{3em} & \hspace{3em}F1\hspace{3em}  \\
\midrule
J48-BR	& 	4.80	(5)	& 	5.90	(8)	& 	4.50	(5)	\\
J48-CC	& 	5.20	(6)	& 	4.25	(4)	& 	4.30	(4)	\\
J48-LP	& 	7.00	(8)	& 	4.95	(5)	& 	6.30	(8)	\\
\midrule
RF-BR	& 	3.20	(2)	& 	5.10	(6)	& 	5.20	(7)	\\
RF-CC	& 	3.40	(3)	& 	4.10	(3)	& 	5.00	(6)	\\
RF-LP	& 	4.30	(4)	& 	2.50	(1)	& 	3.40	(1.5)	\\
\midrule
RDT-DCC	& 	5.40	(7)	& 	5.55	(7)	& 	3.90	(3)	\\
XDCC$_{cum}$	& 	2.70	(1)	& 	3.65	(2)	& 	3.40	(1.5)	\\
\bottomrule
    \end{tabular}


\end{table}

\section{Conclusions}

In this paper, we have shown that the static order of labels is a severe disadvantage of chain-based multi-label classifiers, and have proposed tree-based solutions to overcome this problem.
This is achieved by dynamically selecting the next label in the sequence depending on the context, namely the instance at hand and the previously  predicted labels for it.
In comparison to other approaches for classifier chains, which have to learn appropriate sequences at training time,
our first proposed approach comes at no additional cost, since the framework of random decision trees allows to perform the necessary inferences during prediction time.
%
This also allowed us to confirm the importance of the dynamic label ordering on different datasets
in a controlled setting, where identical random decision tree models were used for static and dynamic chain predictions, so that the observed advantage for the latter can be exclusively attributed to the dynamic label selection.

We have further proposed XDCC, an adaptation of extreme gradient boosted trees to dynamic classifier chains.
It was shown that the positive labels are predominantly predicted at the beginning of the process, which allows XDCC to achieve its maximum  performance already after a few rounds.
This allows XDCC to reduce the length of the chain, which together with the multi-target formulation of XDCC leads to substantial performance improvements in comparison to binary relevance and classifier chains.
The length of the chain also trades off between the two orthogonal objectives of binary relevance and classifier chains, leading to in average the best results in terms of F1.

A key limitation of our approach is that although the above results show that the process reaches optimal performance after a few iterations, we have not thoroughly investigated stopping criteria that would allow an early termination of that process.
To that end, we plan to include a virtual label which indicates the end of the training and prediction process, similar to the idea of the calibrating label in pairwise learning \citep{jf:Neurocomputing}.
This will also help us to address problems with
very large number of labels, which can be further facilitated by integrating some of the sparse techniques proposed by \citet{pmlr-v70-si17a} and \citet{zhang2019GBDT-MO}.
Since the number of associated labels per instance is usually not affected by the increasing number of labels, it will be interesting to see how XDCC will behave with respect to computational costs,
since the size of the (dependency) chains should not grow significantly.
Furthermore, we plan to transfer our ideas on dynamic chains to other kinds of algorithms, such as
predictive clustering trees \citep{vens2008decision}.  
Like random decision trees, their construction also does not
depend on a specific target and is efficient, but they employ
clustering which might yield more discriminative distributions at the leaves.


\begin{footnotesize}
\section*{Declarations}
\begin{description}
\item[\it Funding] n.a.
\item[\it Conflicts of interest] n.a.
\item[\it Ethics approval] n.a.
\item[\it Consent to participate] n.a.
\item[\it Consent for publication] n.a.
\item[\it Availability of data and material] The datasets we used can be found at \url{http://mulan.sf.net/datasets-mlc.html} and \url{http://www.uco.es/kdis/mllresources/}.
\item[\it Code availability]The code of the XDCC algorithm is available at \url{https://github.com/keelm/XDCC}.
\item[\it Authors' contributions]  Eneldo Loza Menc\'\i a conceived the basic idea behind the proposed algorithms and led the implementation, the experimental evaluation, and the write-up. Moritz Kulessa and Simon Bohlender implemented algorithms and conducted the experiments. All authors contributed to the interpretation of the results and the write-up of the paper.
\end{description}
\end{footnotesize}

%
%
%
\footnotesize
\bibliographystyle{splncsnat}
\bibliography{references2,bib,mlc}
%


\newpage
\appendix
\section{Appendix}
\begin{figure}[htb!]
    \centering
    \includegraphics[width=\textwidth]{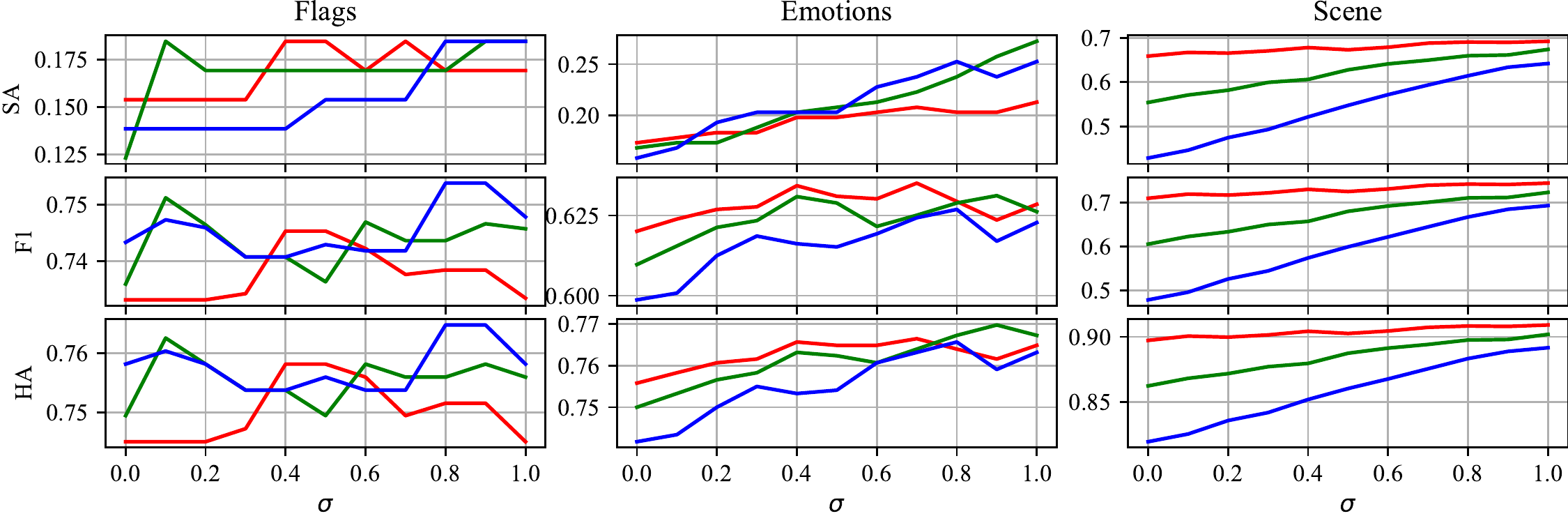}
    \includegraphics[width=\textwidth]{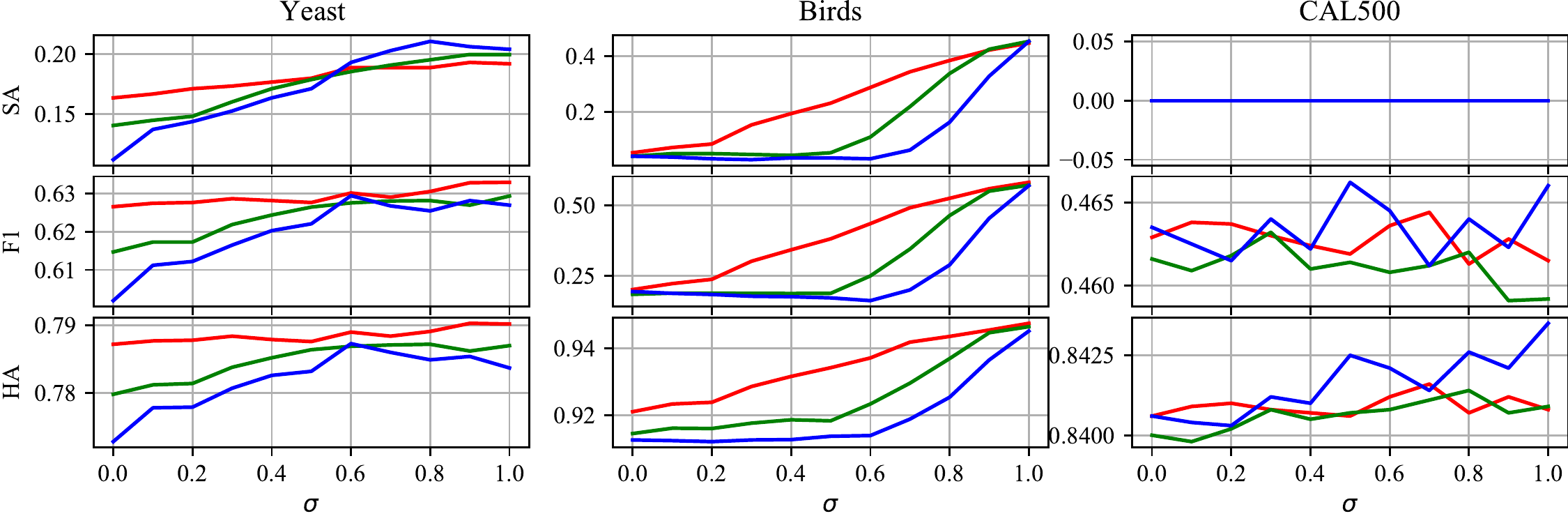}
    \includegraphics[width=\textwidth]{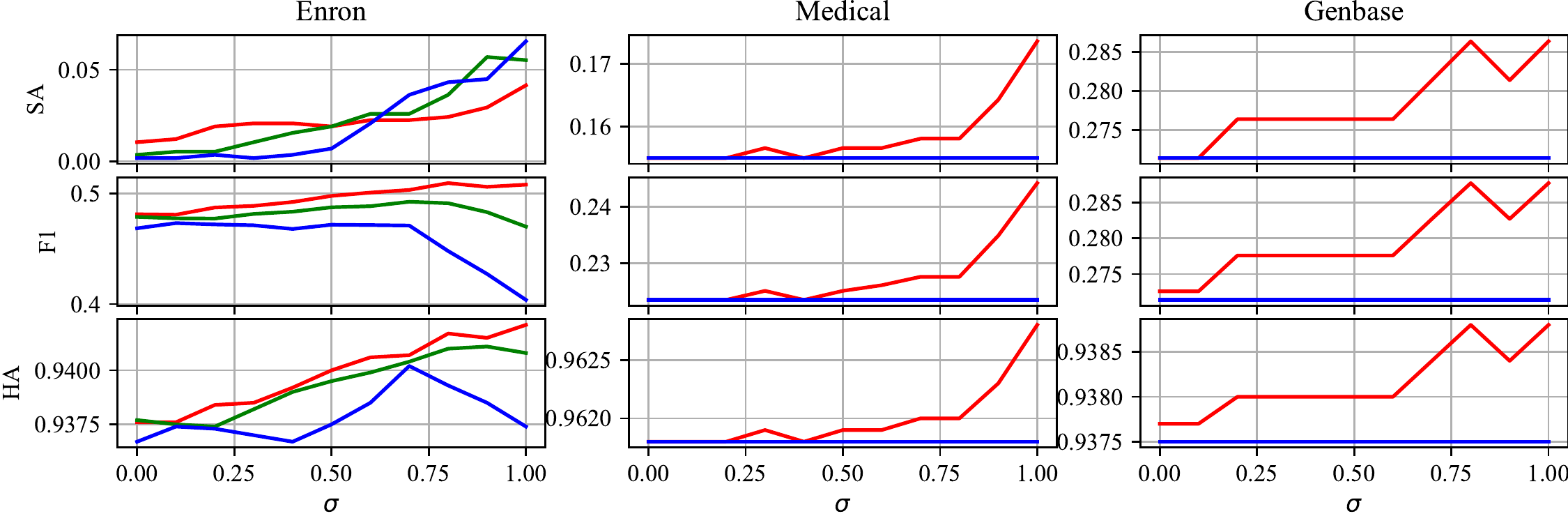}
    \includegraphics[width=0.66\textwidth]{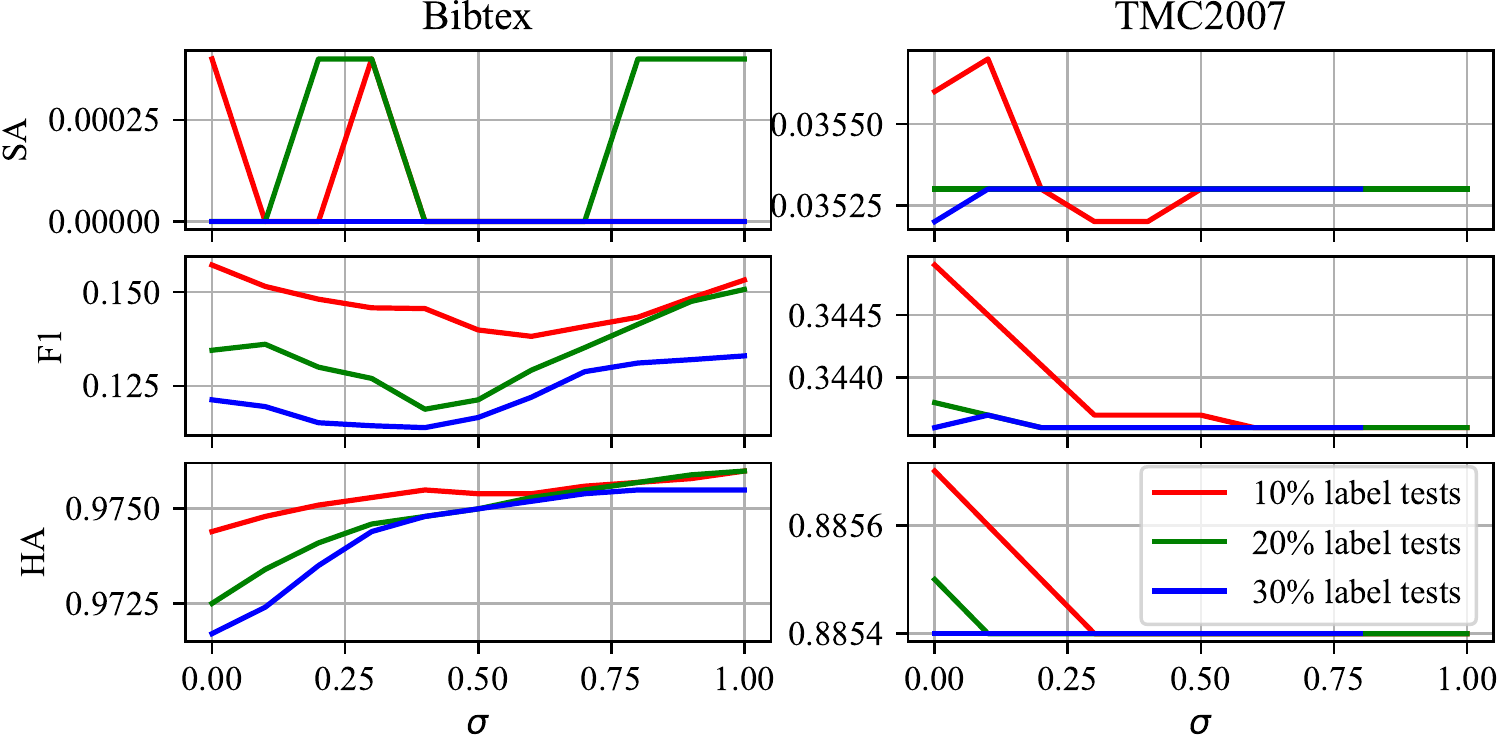}
    \caption{Graphs for all datasets for the experiment described in Section~\ref{sec:rdt_influence_label_tests}}
    \label{fig:rdt_influence_label_tests_appendix}
\end{figure}

The tables and figures in the appendix extend the results shown previously for more datasets or for more algorithms.
Figure~\ref{fig:rdt_influence_label_tests_appendix} shows the influence of the activated label tests in the RDTs on all available datasets (cf. Section~\ref{sec:rdt_influence_label_tests}).
Tables~\ref{tab:ha_results}, \ref{tab:sa_results} and \ref{tab:f1_new} show the performances for all algorithms on all datasets.
ML-RDT denotes the RDT variant for multi-label classification introduced in Section~\ref{sec:RDT}, RDT-LP predicts label sets in the leafs (label powerset transformation).

\begin{table}[htb!]
\begin{tabular}{l|llll}
\toprule
    Algorithm &                \ds{bibtex} &                 \ds{birds} &                \ds{CAL500} &              \ds{emotions} \\
\midrule
       J48-BR &           0.985 (8.0) &          0.949 (11.0) &  0.834 (10.0) &          0.740 (13.0) \\
       J48-CC &           0.985 (9.0) &          0.949 (12.0) &  0.824 (11.0) &          0.751 (11.0) \\
       J48-LP &          0.979 (12.0) &          0.943 (14.0) &  0.805 (12.0) &          0.701 (14.0) \\
\midrule
        RF-BR &           0.986 (6.0) &           0.960 (6.0) &   0.859 (6.0) &           0.799 (2.0) \\
        RF-CC &           0.986 (7.0) &           0.959 (7.0) &   0.859 (5.0) &  \textbf{0.805 (1.0)} \\
        RF-LP &          0.981 (10.0) &           0.961 (3.0) &  0.804 (13.0) &           0.798 (3.0) \\
\midrule
       ML-RDT &          0.979 (13.0) &          0.946 (13.0) &   0.841 (9.0) &          0.748 (12.0) \\
       RDT-LP &          0.981 (11.0) &           0.956 (9.0) &  0.796 (14.0) &           0.795 (4.0) \\
\midrule
       XGB-BR &  \textbf{0.988 (1.0)} &  \textbf{0.963 (1.0)} &   0.862 (3.0) &          0.773 (10.0) \\
       XGB-CC &           0.987 (2.0) &           0.963 (2.0) &   0.854 (7.0) &           0.793 (6.0) \\
\midrule
      RDT-DCC &          0.977 (14.0) &          0.950 (10.0) &   0.842 (8.0) &           0.776 (8.0) \\
       ML-XGB &           0.987 (4.0) &           0.961 (4.0) &   \textbf{0.864 (1.5)} &           0.782 (7.0) \\
 $XDCC_{cum}$ &           0.987 (3.0) &           0.960 (5.0) &   \textbf{0.864 (1.5)} &           0.794 (5.0) \\
 $XDCC_{std}$ &           0.987 (5.0) &           0.958 (8.0) &   0.860 (4.0) &           0.776 (9.0) \\
\bottomrule
\multicolumn{5}{c}{} \\[0.2cm]
\toprule
    Algorithm &                 \ds{enron} &                 \ds{flags} &               \ds{genbase} &               \ds{medical} \\
\midrule
       J48-BR &           0.946 (9.0) &   0.725 (8.0) &           0.999 (3.0) &           0.989 (2.0) \\
       J48-CC &           0.948 (8.0) &  0.701 (12.0) &           0.999 (3.0) &  \textbf{0.990 (1.0)} \\
       J48-LP &          0.927 (14.0) &  0.697 (13.0) &           0.999 (5.0) &           0.983 (8.0) \\
\midrule
        RF-BR &           0.953 (4.0) &   0.743 (6.0) &           0.997 (9.0) &          0.980 (11.0) \\
        RF-CC &           0.952 (5.0) &   0.734 (7.0) &          0.997 (10.0) &          0.980 (10.0) \\
        RF-LP &          0.942 (11.0) &   0.710 (9.0) &           0.999 (3.0) &           0.983 (9.0) \\
\midrule
       ML-RDT &          0.941 (12.0) &   0.754 (3.0) &          0.943 (12.0) &          0.965 (13.0) \\
       RDT-LP &          0.940 (13.0) &  0.706 (10.0) &          0.939 (14.0) &          0.970 (12.0) \\
\midrule
       XGB-BR &  \textbf{0.953 (1.0)} &   \textbf{0.774 (1.5)} &           0.998 (6.0) &           0.989 (3.5) \\
       XGB-CC &           0.952 (7.0) &  0.690 (14.0) &  \textbf{0.999 (1.0)} &           0.989 (3.5) \\
\midrule
      RDT-DCC &          0.942 (10.0) &   0.750 (4.0) &          0.941 (13.0) &          0.964 (14.0) \\
       ML-XGB &           0.953 (3.0) &   \textbf{0.774 (1.5)} &           0.998 (7.5) &           0.989 (6.0) \\
 $XDCC_{cum}$ &           0.953 (2.0) &   0.745 (5.0) &           0.998 (7.5) &           0.989 (5.0) \\
 $XDCC_{std}$ &           0.952 (6.0) &  0.703 (11.0) &          0.991 (11.0) &           0.988 (7.0) \\
\bottomrule
\multicolumn{5}{c}{} \\[0.2cm]
\toprule
    Algorithm &                 \ds{scene} &               \ds{tmc2007} &                 \ds{yeast} \\
\midrule
       J48-BR &          0.861 (12.0) &             - &          0.741 (12.0) \\
       J48-CC &          0.857 (13.0) &             - &          0.726 (13.0) \\
       J48-LP &          0.852 (14.0) &             - &          0.710 (14.0) \\
\midrule
        RF-BR &           0.911 (7.0) &             - &  \textbf{0.806 (1.0)} \\
        RF-CC &           0.915 (4.0) &             - &           0.801 (3.0) \\
        RF-LP &  \textbf{0.918 (1.0)} &             - &           0.795 (6.0) \\
\midrule
       ML-RDT &           0.916 (3.0) &           0.885 (7.5) &           0.790 (7.0) \\
       RDT-LP &           0.913 (5.0) &           0.889 (6.0) &          0.785 (10.0) \\
\midrule
       XGB-BR &           0.917 (2.0) &           0.934 (4.0) &           0.801 (2.0) \\
       XGB-CC &           0.912 (6.0) &           0.934 (2.0) &           0.785 (8.0) \\
\midrule
      RDT-DCC &           0.904 (8.0) &           0.885 (7.5) &           0.785 (9.0) \\
       ML-XGB &          0.896 (10.5) &  \textbf{0.935 (1.0)} &           0.799 (5.0) \\
 $XDCC_{cum}$ &          0.896 (10.5) &           0.933 (5.0) &           0.800 (4.0) \\
 $XDCC_{std}$ &           0.896 (9.0) &           0.934 (3.0) &          0.782 (11.0) \\
\bottomrule
\end{tabular}
    \caption{Hamming accuracy  results of all algorithms on all datasets. Ranks are shown in brackets.}
    \label{tab:ha_results}
\end{table}

\begin{table}[]
\begin{tabular}{l|llll}
\toprule
    Algorithm &                \ds{bibtex} &                 \ds{birds} &                \ds{CAL500} &              \ds{emotions} \\
\midrule
       J48-BR &           0.133 (8.0) &          0.486 (10.5) &  0.000 (7.5) &          0.129 (14.0) \\
       J48-CC &           0.144 (4.0) &          0.486 (10.5) &  0.000 (7.5) &          0.213 (10.0) \\
       J48-LP &           0.143 (5.0) &          0.468 (13.0) &  0.000 (7.5) &           0.218 (9.0) \\
\midrule
        RF-BR &          0.084 (11.0) &           0.523 (7.0) &  0.000 (7.5) &           0.257 (7.0) \\
        RF-CC &          0.083 (12.0) &           0.517 (8.0) &  0.000 (7.5) &           0.292 (3.0) \\
        RF-LP &           0.144 (3.0) &           0.532 (5.0) &  0.000 (7.5) &  \textbf{0.376 (1.0)} \\
\midrule
       ML-RDT &          0.020 (13.0) &          0.371 (14.0) &  0.000 (7.5) &          0.178 (13.0) \\
       RDT-LP &          0.132 (10.0) &           0.495 (9.0) &  0.000 (7.5) &           0.361 (2.0) \\
\midrule
       XGB-BR &  \textbf{0.170 (1.0)} &  \textbf{0.576 (1.0)} &  0.000 (7.5) &          0.198 (11.0) \\
       XGB-CC &           0.169 (2.0) &           0.560 (2.0) &  0.000 (7.5) &           0.287 (4.0) \\
\midrule
      RDT-DCC &          0.009 (14.0) &          0.480 (12.0) &  0.000 (7.5) &           0.274 (5.0) \\
       ML-XGB &           0.142 (6.0) &           0.539 (3.0) &  0.000 (7.5) &           0.228 (8.0) \\
 $XDCC_{cum}$ &           0.140 (7.0) &           0.536 (4.0) &  0.000 (7.5) &           0.267 (6.0) \\
 $XDCC_{std}$ &           0.132 (9.0) &           0.523 (6.0) &  0.000 (7.5) &          0.188 (12.0) \\
\bottomrule
\multicolumn{5}{c}{} \\[0.2cm]
\toprule
    Algorithm &                 \ds{enron} &                 \ds{flags} &               \ds{genbase} &               \ds{medical} \\
\midrule
       J48-BR &          0.086 (12.0) &          0.077 (14.0) &           0.975 (3.0) &           0.651 (3.0) \\
       J48-CC &           0.116 (9.0) &           0.185 (4.5) &           0.975 (3.0) &  \textbf{0.682 (1.0)} \\
       J48-LP &          0.095 (11.0) &           0.200 (2.5) &           0.970 (5.0) &           0.594 (8.0) \\
\midrule
        RF-BR &           0.124 (5.0) &           0.169 (9.5) &           0.940 (9.0) &          0.281 (12.0) \\
        RF-CC &           0.133 (4.0) &           0.185 (4.5) &          0.935 (10.0) &          0.326 (10.0) \\
        RF-LP &  \textbf{0.173 (1.0)} &           0.169 (9.5) &           0.975 (3.0) &           0.555 (9.0) \\
\midrule
       ML-RDT &          0.031 (14.0) &          0.154 (11.5) &          0.312 (12.0) &          0.219 (13.0) \\
       RDT-LP &           0.161 (2.0) &          0.154 (11.5) &          0.296 (14.0) &          0.295 (11.0) \\
\midrule
       XGB-BR &           0.116 (8.0) &           0.200 (2.5) &           0.960 (7.0) &           0.637 (7.0) \\
       XGB-CC &           0.136 (3.0) &           0.169 (7.5) &  \textbf{0.980 (1.0)} &           0.640 (6.0) \\
\midrule
      RDT-DCC &          0.047 (13.0) &           0.172 (6.0) &          0.301 (13.0) &          0.207 (14.0) \\
       ML-XGB &           0.119 (7.0) &  \textbf{0.246 (1.0)} &           0.960 (7.0) &           0.648 (4.0) \\
 $XDCC_{cum}$ &           0.121 (6.0) &           0.169 (7.5) &           0.960 (7.0) &           0.654 (2.0) \\
 $XDCC_{std}$ &          0.105 (10.0) &          0.123 (13.0) &          0.874 (11.0) &           0.643 (5.0) \\
\bottomrule
\multicolumn{5}{c}{} \\[0.2cm]
\toprule
    Algorithm &                 \ds{scene} &               \ds{tmc2007} &                 \ds{yeast} \\
\midrule
       J48-BR &          0.401 (14.0) &             - &          0.064 (13.0) \\
       J48-CC &          0.530 (13.0) &             - &          0.121 (11.0) \\
       J48-LP &          0.536 (12.0) &             - &          0.117 (12.0) \\
\midrule
        RF-BR &          0.550 (11.0) &             - &           0.163 (9.0) \\
        RF-CC &          0.569 (10.0) &             - &           0.205 (3.0) \\
        RF-LP &  \textbf{0.722 (1.0)} &             - &  \textbf{0.257 (1.0)} \\
\midrule
       ML-RDT &           0.714 (2.0) &           0.036 (7.0) &           0.178 (7.0) \\
       RDT-LP &           0.708 (3.0) &           0.086 (6.0) &           0.244 (2.0) \\
\midrule
       XGB-BR &           0.584 (9.0) &           0.257 (2.0) &           0.183 (6.0) \\
       XGB-CC &           0.611 (6.0) &  \textbf{0.262 (1.0)} &          0.136 (10.0) \\
\midrule
      RDT-DCC &           0.678 (4.0) &           0.035 (8.0) &           0.201 (4.0) \\
       ML-XGB &           0.595 (7.5) &           0.249 (4.0) &           0.177 (8.0) \\
 $XDCC_{cum}$ &           0.595 (7.5) &           0.250 (3.0) &           0.200 (5.0) \\
 $XDCC_{std}$ &           0.625 (5.0) &           0.238 (5.0) &          0.047 (14.0) \\
\bottomrule
\end{tabular}
    \caption{Subset accuracy  results of all algorithms on all datasets. Ranks are shown in brackets.}
    \label{tab:sa_results}
\end{table}

\begin{table}[b]
\begin{tabular}{l|llll}
\toprule
    Algorithm &                \ds{bibtex} &                 \ds{birds} &                \ds{CAL500} &              \ds{emotions} \\
\midrule
       J48-BR &           0.366 (2.0) &           0.603 (6.0) &           0.342 (5.0) &          0.540 (12.0) \\
       J48-CC &           0.347 (4.0) &           0.598 (9.0) &           0.353 (3.0) &           0.560 (9.0) \\
       J48-LP &           0.303 (5.0) &          0.597 (10.0) &           0.331 (8.0) &          0.498 (14.0) \\
\midrule
        RF-BR &          0.176 (12.0) &          0.594 (11.0) &           0.346 (4.0) &           0.581 (8.0) \\
        RF-CC &          0.174 (13.0) &          0.594 (12.0) &           0.341 (6.0) &           0.636 (4.0) \\
        RF-LP &          0.237 (10.0) &           0.668 (3.0) &           0.333 (7.0) &  \textbf{0.669 (1.0)} \\
\midrule
       ML-RDT &           0.262 (9.0) &          0.549 (13.0) &  \textbf{0.465 (1.0)} &           0.608 (5.0) \\
       RDT-LP &          0.164 (14.0) &          0.543 (14.0) &           0.325 (9.0) &           0.668 (2.0) \\
\midrule
       XGB-BR &  \textbf{0.372 (1.0)} &  \textbf{0.676 (1.0)} &          0.320 (12.0) &          0.545 (11.0) \\
       XGB-CC &           0.358 (3.0) &           0.670 (2.0) &          0.233 (13.0) &           0.608 (6.0) \\
\midrule
      RDT-DCC &          0.190 (11.0) &           0.600 (8.0) &           0.463 (2.0) &           0.641 (3.0) \\
       ML-XGB &           0.296 (7.0) &           0.614 (5.0) &          0.324 (10.5) &          0.558 (10.0) \\
 $XDCC_{cum}$ &           0.301 (6.0) &           0.614 (4.0) &          0.324 (10.5) &           0.583 (7.0) \\
 $XDCC_{std}$ &           0.290 (8.0) &           0.600 (7.0) &          0.160 (14.0) &          0.531 (13.0) \\
\bottomrule
\multicolumn{5}{c}{} \\[0.2cm]
\toprule
    Algorithm &                 \ds{enron} &                 \ds{flags} &               \ds{genbase} &               \ds{medical} \\
\midrule
       J48-BR &          0.473 (11.0) &           0.711 (8.0) &           0.991 (3.0) &           0.773 (2.0) \\
       J48-CC &           0.503 (9.0) &          0.659 (12.0) &           0.991 (3.0) &  \textbf{0.777 (1.0)} \\
       J48-LP &          0.411 (14.0) &          0.668 (11.0) &           0.988 (8.0) &           0.701 (8.0) \\
\midrule
        RF-BR &           0.505 (7.0) &           0.724 (6.0) &           0.966 (9.0) &          0.353 (12.0) \\
        RF-CC &           0.525 (2.0) &           0.715 (7.0) &          0.957 (10.0) &          0.397 (10.0) \\
        RF-LP &          0.486 (10.0) &          0.683 (10.0) &           0.991 (3.0) &           0.676 (9.0) \\
\midrule
       ML-RDT &           0.515 (5.0) &           0.741 (2.0) &          0.334 (12.0) &          0.299 (13.0) \\
       RDT-LP &          0.429 (13.0) &           0.692 (9.0) &          0.296 (14.0) &          0.394 (11.0) \\
\midrule
       XGB-BR &           0.521 (3.0) &  \textbf{0.758 (1.0)} &           0.989 (5.0) &           0.753 (5.0) \\
       XGB-CC &  \textbf{0.534 (1.0)} &          0.635 (13.0) &  \textbf{0.992 (1.0)} &           0.747 (7.0) \\
\midrule
      RDT-DCC &           0.505 (8.0) &           0.739 (3.0) &          0.307 (13.0) &          0.279 (14.0) \\
       ML-XGB &           0.509 (6.0) &           0.739 (4.0) &           0.988 (6.5) &           0.753 (4.0) \\
 $XDCC_{cum}$ &           0.518 (4.0) &           0.738 (5.0) &           0.988 (6.5) &           0.758 (3.0) \\
 $XDCC_{std}$ &          0.467 (12.0) &          0.618 (14.0) &          0.943 (11.0) &           0.752 (6.0) \\
\bottomrule
\multicolumn{5}{c}{} \\[0.2cm]
\toprule
    Algorithm &                 \ds{scene} &               \ds{tmc2007} &                 \ds{yeast} \\
\midrule
       J48-BR &          0.552 (14.0) &             - &          0.547 (11.0) \\
       J48-CC &          0.604 (11.0) &             - &          0.528 (12.0) \\
       J48-LP &          0.588 (13.0) &             - &          0.494 (14.0) \\
\midrule
        RF-BR &          0.599 (12.0) &             - &           0.609 (8.0) \\
        RF-CC &          0.608 (10.0) &             - &           0.622 (5.0) \\
        RF-LP &  \textbf{0.773 (1.0)} &             - &           0.632 (2.0) \\
\midrule
       ML-RDT &           0.764 (2.0) &           0.344 (6.0) &  \textbf{0.633 (1.0)} \\
       RDT-LP &           0.756 (3.0) &           0.177 (8.0) &           0.620 (6.0) \\
\midrule
       XGB-BR &           0.643 (9.0) &           0.615 (3.0) &           0.613 (7.0) \\
       XGB-CC &           0.664 (8.0) &           0.622 (2.0) &          0.559 (10.0) \\
\midrule
      RDT-DCC &           0.729 (4.0) &           0.344 (7.0) &           0.629 (4.0) \\
       ML-XGB &           0.686 (5.5) &           0.600 (4.0) &           0.602 (9.0) \\
 $XDCC_{cum}$ &           0.686 (5.5) &  \textbf{0.624 (1.0)} &           0.631 (3.0) \\
 $XDCC_{std}$ &           0.673 (7.0) &           0.590 (5.0) &          0.507 (13.0) \\
\bottomrule
\end{tabular}

    \caption{F1 results of all algorithms on all datasets. Ranks are shown in brackets.}
    \label{tab:f1_new}
\end{table}

\raus{
\begin{table}[tb]
  \caption{Predictive performance and training times comparison. Shown are the average ranks over the 10 datasets and the ranks over these in brackets.\tod{If time, make ranks of ranks smaller and put more space between columns.} }
  \label{tab:overallcomparison_all}
\centering
\raus{
\scalebox{0.91}{
    \begin{tabular}{lcccc}
\toprule
Method & \hspace{2em}HA\hspace{2em} & \hspace{2em}SA\hspace{2em} & \hspace{2em}F1\hspace{2em} & \hspace{0em}Train time\hspace{0em}  \\
\midrule
J48-BR	& 	9.90	(10)	& 	10.70	(13)	& 	7.80	(8)	& 	8.70	(9)	\\
J48-CC	& 	10.30	(12)	& 	7.55	(8)	& 	7.60	(7)	& 	8.30	(8)	\\
J48-LP	& 	13.30	(16)	& 	9.00	(9)	& 	11.00	(13)	& 	5.10	(5)	\\
\midrule
RF-BR	& 	6.30	(4.5)	& 	9.15	(10)	& 	9.10	(11)	& 	10.40	(11)	\\
RF-CC	& 	6.30	(4.5)	& 	7.30	(6)	& 	8.10	(10)	& 	9.00	(10)	\\
RF-LP	& 	7.70	(7)	& 	4.35	(1)	& 	5.80	(3)	& 	7.80	(7)	\\
\midrule
RDT-PM	& 	9.95	(11)	& 	13.80	(16)	& 	13.90	(16)	& 	2.05	(1)	\\
RDT-LP	& 	11.70	(14)	& 	7.50	(7)	& 	9.90	(12)	& 	2.40	(3)	\\
RDT-LM	& 	11.10	(13)	& 	11.50	(15)	& 	6.40	(4)	& 	2.35	(2)	\\
\midrule
XGB-BR	& 	3.40	(1)	& 	6.10	(4)	& 	5.60	(1)	& 	13.80	(13)	\\
XGB-CC	& 	6.35	(6)	& 	5.20	(2)	& 	6.90	(5)	& 	14.60	(14)	\\
RDT-DCC-LM	& 	12.05	(15)	& 	10.65	(12)	& 	8.05	(9)	& 	4.80	(4)	\\
RDT-DCC-PM	& 	8.50	(8)	& 	11.30	(14)	& 	12.15	(15)	& 	5.40	(6)	\\
\midrule
ML-XGB	& 	5.20	(2)	& 	6.00	(3)	& 	6.95	(6)	& 	11.90	(12)	\\
XDCC$_{cum}$	& 	5.25	(3)	& 	6.15	(5)	& 	5.65	(2)	& 	14.70	(15.5)	\\
XDCC$_{std}$	& 	8.70	(9)	& 	9.75	(11)	& 	11.10	(14)	& 	14.70	(15.5)	\\
\bottomrule
    \end{tabular}}
}
\end{table}
}

\end{document}